\documentclass[10pt]{article} 

\usepackage[hidelinks]{hyperref}
\usepackage{enumitem}
\usepackage{tabularx}
\usepackage{adjustbox}
\usepackage{comment}
\usepackage{geometry} 
\usepackage{comment}

\usepackage{authblk} 

\usepackage[accepted]{tmlr}

\usepackage{amsmath,amsfonts,bm}

\def\eqref#1{equation~\ref{#1}}

\def\1{\bm{1}}

\DeclareMathAlphabet{\mathsfit}{\encodingdefault}{\sfdefault}{m}{sl}
\SetMathAlphabet{\mathsfit}{bold}{\encodingdefault}{\sfdefault}{bx}{n}

\usepackage{url}

\usepackage{microtype}
\usepackage{graphicx}
\usepackage{booktabs} 

\usepackage{algorithm}
\let\classAND\AND
\let\AND\relax
\usepackage{algorithmic}

\let\AND\classAND
\AtBeginEnvironment{algorithmic}{\let\AND\algoAND}
\usepackage{amsmath}

\usepackage{amssymb}
\usepackage{mathtools}
\usepackage{amsthm}
\usepackage{multirow}
\usepackage{adjustbox}
\usepackage{booktabs} 
\usepackage{siunitx} 
\usepackage[normalem]{ulem}
\usepackage{subfig}
\usepackage{float}

\usepackage{hyperref}    
\hypersetup{
    colorlinks=true,          
    linkcolor=blue,           
    filecolor=blue,        
    urlcolor=blue,            
    citecolor=blue,          
}

\usepackage[capitalize,noabbrev]{cleveref}

\theoremstyle{plain}

\theoremstyle{definition}

\theoremstyle{remark}

\title{\centering  \textbf{\texttt{Yan}: Foundational Interactive Video Generation}}

\date{}

\author{\normalsize Yan Team}

\affil{Tencent}
\begin{document}

\newgeometry{top=0.3in, bottom=1in, left=1in, right=1in}
\maketitle

\begin{figure}[h]
\centering
\includegraphics[width=1\textwidth]{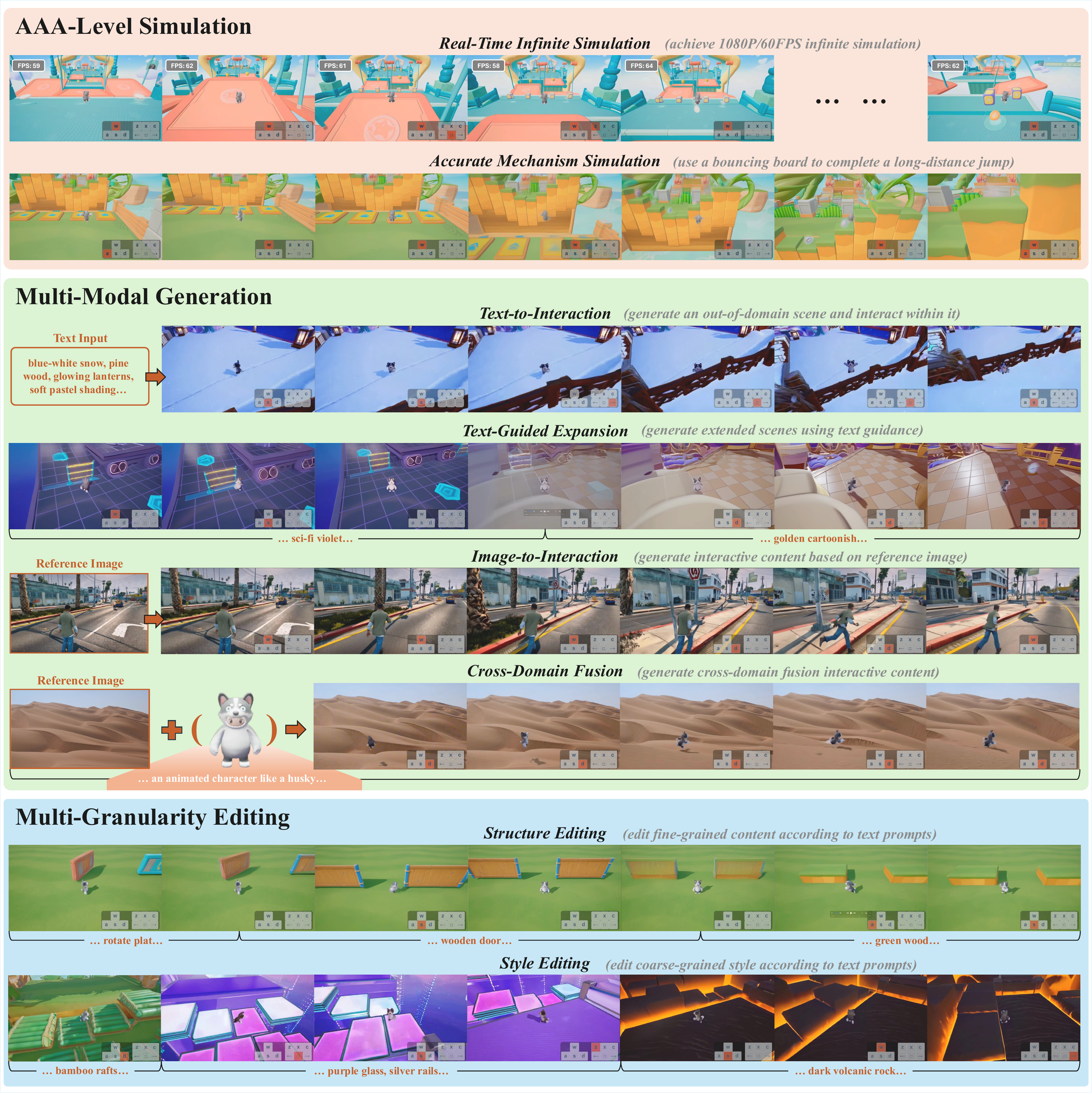}

\caption{
\textbf{Comprehensive capabilities of \texttt{Yan}}. 
\texttt{Yan} supports real-time interactive video generation, with all interactions driven by user input. It offers a wide range of capabilities, including AAA-level simulations, multi-modal generation, and multi-granularity editing. Notably, during the editing process, users can dynamically modify prompts to edit subsequent generated content interactively.
}

\vspace{2em}
\label{fig:pr}
\end{figure}

\clearpage
\restoregeometry 

\begin{abstract}

We present \texttt{Yan}, a foundational framework for interactive video generation, covering the entire pipeline from simulation and generation to editing. Specifically, \texttt{Yan} comprises three core modules. \textit{AAA-level Simulation}: We design a highly-compressed, low-latency 3D-VAE coupled with a KV-cache-based shift-window denoising inference process, achieving real-time 1080P/60FPS interactive simulation. \textit{Multi-Modal Generation}: We introduce a hierarchical autoregressive caption method that injects game-specific knowledge into open-domain multi-modal video diffusion models (VDMs), then transforming the VDM into a frame-wise, action-controllable, real-time infinite interactive video generator. Notably, when the textual and visual prompts are sourced from different domains, the model demonstrates strong generalization, allowing it to blend and compose the style and mechanics across domains flexibly according to user prompts. \textit{Multi-Granularity Editing}: We propose a hybrid model that explicitly disentangles interactive mechanics simulation from visual rendering, enabling multi-granularity video content editing during interaction through text. Collectively, \texttt{Yan} offers an integration of these modules, pushing interactive video generation beyond isolated capabilities toward a comprehensive AI-driven interactive creation paradigm, paving the way for the next generation of creative tools, media, and entertainment.
The project is shown at:  \href{https://greatx3.github.io/Yan/}{here}.

\end{abstract}

\section{Introduction}

Recent breakthroughs in Artificial Intelligence Generated Content (AIGC) have evolved from text~\citep{aga,ma,llm_survey} and images~\citep{stable_diffusion} to video synthesis~\citep {sora}, now reaching a transformative frontier: interactive generative video (IGV)~\citep{igv_survey}. Unlike conventional video generation, IGV dynamically responds to user inputs, enabling real-time, personalized visual experiences. This paradigm holds the potential to revolutionize fields that demand ongoing human-AI interaction, ranging from immersive media and virtual simulation to embodied intelligence. However, it confronts significant challenges: existing methods ~\citep{diamond,genie,genie2} struggle to simultaneously attain high visual fidelity, sustained temporal coherence, and rich interactivity. Moreover, generated content typically remains static post-creation, lacking real-time adaptability or user customization.

The challenges manifest acutely in complex interactive scenarios. Consider modern video games: they demand not only 1080P/60FPS rendering and intricate physics simulation, but also granular response to diverse inputs and dynamic content adaptation. These requirements make games an ideal testbed for IGV capabilities. However, existing methods remain with critical gaps. For instance, GameNGen~\citep{gameNgen}, PlayGen~\citep{playgen}, and MineWorld~\citep{mineworld} suffer from limited visual quality and generalizability. Meanwhile, The-Matrix~\citep{the_matrix}, GameFactory~\citep{gamefactory}, and Matrix-Game~\citep{matrix_game} lack intricate physics simulation and fall short of real-time performance. Besides, these methods treat interactive video as fixed content generation, unable to support dynamic modification during interaction, severely restricting practical creativity and flexibility. Thus, three core challenges remain unresolved: (1) achieving high-fidelity, real-time visual experience; (2) enabling generalizable and prompt-controllable generation; and (3) empowering dynamic, interactive editing and on-the-fly content customization during interaction.

To tackle these challenges, we propose \texttt{Yan}, a foundational framework for interactive video generation, integrating three targeted modules trained on a shared dataset from modern 3D game environments~\footnote{\url{https://en.wikipedia.org/wiki/Yuan_Meng_Star}}. Specifically, for the first challenge of real-time, high-fidelity visual experience, our \textit{AAA-level Simulation} module leverages a high-compression, low-latency 3D-VAE together with KV-cache-based shift-window denoising inference process, achieving unprecedented performance at 1080P/60FPS while preserving complex mechanics. To address the second challenge of prompt-controllable and generalizable content generation, our \textit{multi-modal generation} module introduces a hierarchical captioning method that injects game-specific knowledge into open-domain video diffusion models, and further transforms them for frame-wise, action-controllable generation. This supports real-time interactive content creation driven by both text and image prompts across diverse domains. Finally, to tackle interactive editing, our \textit{multi-granularity editing} module adopts a hybrid architecture that explicitly disentangles interactive mechanics simulation from visual rendering, supporting coarse-to-fine control and real-time editing of video content at any moment during interaction. The comprehensive capabilities of \texttt{Yan} are illustrated in Fig.~\ref{fig:pr}.

We believe \texttt{Yan} marks a milestone---moving interactive video generation from fragmented prototypes toward a coherent, and truly generative paradigm, and charting a path for the next generation of open, creative, AI-powered digital worlds.

\section{Related Work}

\paragraph{Interactive Generative Video}



Prior work falls into two lines. (i) Game-centric interactive video generation transfers action control and structural priors learned from game datasets to open-ended settings ~\citep{gamefactory,mineworld,matrix_game,the_matrix,genie,genie2}. These methods typically rely on action annotations and demonstrate some generalization beyond the source game, but they are often not truly frame-wise interactive (chunked control or high latency), emphasize navigation over accurate physics, and run at modest resolutions and real-time capability. (ii) Real-world world models aim to predict future frames or trajectories under control~\citep{nwm,unisim,cosmos}, but they generally do not provide real-time, frame-wise action responsiveness with promptable content control. In contrast, Yan targets real-time, causal, frame-wise control with accurate physics, high visual fidelity, and prompt-controllable generation in open domains.

\paragraph{Game Simulation with Neural Networks}
Works in this category aim to simulate a game using neural networks. MarioVGG \citep{mariovgg} generates game video segments of \emph{Super Mario Bros} \citep{mario_dataset} from text-based actions and the first frame of the game, but it lacks real-time capabilities. GameNGen \citep{gameNgen} uses a diffusion model to simulate games at 20 FPS on \emph{Doom} \citep{vizdoom}. PlayGen \citep{playgen} achieves 20 FPS on both \emph{Doom} and \emph{Super Mario Bros} \citep{java_mario}. Oasis \citep{oasis} is able to simulate \emph{Minecraft} \citep{minecraft} at 20 FPS. However, all the above approaches are constrained to relatively low-resolution results and limited frame rates, and are mainly tested in simplified or classic 2D/3D game environments. In contrast, \texttt{Yan} targets a modern 3D game \emph{Yuan Meng Star} \citep{ymzx}, achieving 1080P rendering at 60 FPS while preserving intricate game mechanics, expanding the capabilities and application range of game simulation. 

\paragraph{Video Editing.}

Diffusion-based video editing methods transform a given source video while preserving spatiotemporal coherence ~\citep{sora,wan,hunyuanvideo,pix2video,video_p2p,vace}. However, these approaches assume non-interactive inputs and do not ensure that edited content remains responsive to user actions in real time. Within interactive video generation, most efforts prioritize high-fidelity simulation~\citep{gameNgen,oasis,playgen,matrix_game}, leaving interactive video editing largely unaddressed. Yan fills this gap by explicitly disentangling mechanics simulation from visual rendering: a depth-driven mechanics simulator preserves structure-dependent physics and interactivity, while a renderer—guided by textual prompts—handles style. This design enables on-the-fly, multi-granularity edits (both structure and style) during interaction, with temporal consistency and real-time action alignment.

\section{Overview}

\begin{figure}[t]
\centering
\includegraphics[width=0.99\textwidth]{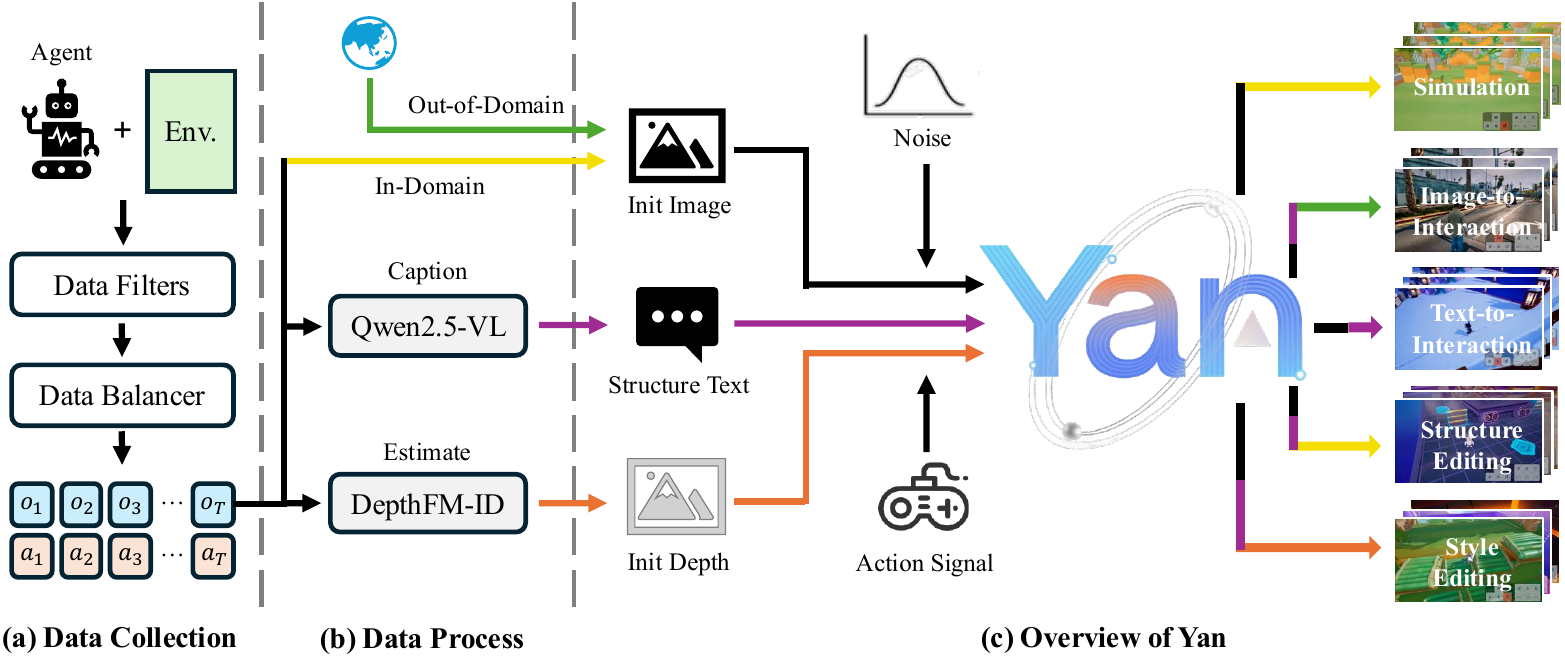}
\caption{
The overall framework of \texttt{Yan}. First, we leverage the agent to collect and clean data in the game environment. Next, we use a vision language model and a depth estimation model to annotate the collected data, generating a structured text prompt and depth. Finally, both the labeled data and the unlabeled open-domain data are used together for training.
}
\label{fig:framework}
\end{figure}

\texttt{Yan} presents a unified framework for interactive video generation, integrating the full pipeline from data collection to simulation, generation, and editing, as illustrated in Fig.~\ref{fig:framework}. The process begins with the automated gathering of a large-scale, high-quality interactive video dataset in a modern 3D game environment, ensuring both diversity and precise action-visual correspondence. At its core, Yan consists of three interconnected modules:
\begin{itemize}
  \item \textbf{Yan-Sim: AAA-level Simulation} achieves high-fidelity, real-time rendering and physics simulation, seamlessly translating user actions into 1080P/60FPS interactive visuals.
  \item \textbf{Yan-Gen: Multi-Modal Generation} enables prompt-controllable and generalizable video generation guided by text or images, supporting creative content formation and style/mechanics blending.
  \item \textbf{Yan-Edit: Multi-Granularity Editing} disentangles interactive mechanics simulation and visual rendering, allowing flexible, real-time editing of both structure and style during interaction through simple text prompts.
\end{itemize}
These modules are deeply integrated to deliver a coherent, end-to-end workflow for truly interactive video generation. Each component will be discussed in detail in the following sections: data collection and preprocessing in Sec.~\ref{sec:data}, simulation (\textbf{Yan-Sim}) in Sec.~\ref{sec:sim}, generation (\textbf{Yan-Gen}) in Sec.~\ref{sec:gen}, and editing (\textbf{Yan-Edit}) in Sec.~\ref{sec:edit}.
\section{Data Collection}
\label{sec:data}

\begin{figure}[t]
\centering
\includegraphics[width=0.99\textwidth]{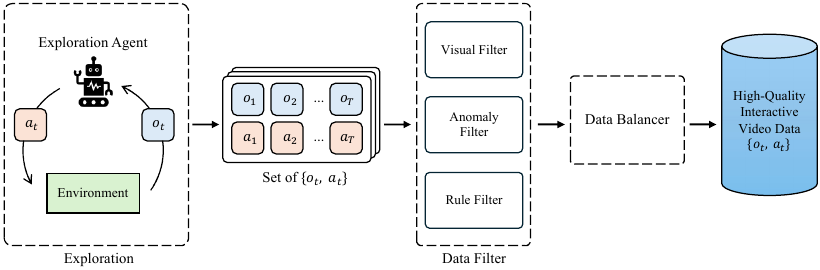}
\caption{
The overview of the data collection pipeline. 
}
\label{fig:data_collection}
\end{figure}


In this section, we present the full pipeline of constructing a high-quality interactive video dataset, which serves as a shared dataset for training \texttt{Yan-Sim}, \texttt{Yan-Gen}, and \texttt{Yan-Edit}. Compared to video datasets~\citep{openvid-1m,internvid,panda70m,goku} used for video generation tasks that do not involve frame-wise interaction, the interactive video datasets should additionally include per-frame interaction annotations to reflect the transition between two adjacent frames of the video (e.g., the movement of the subject in the video). Previous related works~\citep{gameNgen,playgen} usually utilize interactive video datasets that only contain a few scenarios with simple physical laws. To enable interactive video generation with intricate physical laws in various scenarios, we develop an automated interactive video data collection pipeline, as shown in Fig.~\ref{fig:data_collection}. We select a renowned modern 3D game~\citep{ymzx} as our data collection environment due to its complex and diverse scenarios, and rich action space. With this pipeline, we are able to efficiently collect interactive video data, providing a solid foundation for training interactive video generation models. The details of the pipeline are described below.

\subsection{Data Collection Pipeline}

\noindent
\textbf{Exploration Agent.}
Collecting interactive video data manually is tedious and complex, hindering the automation of the collection pipeline. 
%
To address this, we develop a scene exploration agent to automatically interact with the environment to generate interactive video data. 
Specifically, it will take action $a_t$ (i.e., the interaction annotation) to interact with the environment according to the image state $o_t$ of the environment at each timestep $t$.
%
The interaction strategy of this agent is decided by a combination of a random model that chooses random actions at every interaction and a reinforcement learning based (RL) model trained using the proximal policy optimization (PPO)~\citep{ppo} algorithm, similar to PlayGen~\citep{playgen}. 
%
Our data collection environment covers more than 90 scenarios with different interactive mechanics, each presenting unique exploration challenges. 
Using only the random model is able to collect data in a small area around the initial position of each scenario.
Hence, the RL model is employed to enable the agent to explore various positions in each scenario.
%
%
In general, we leverage the random model and the RL model to increase the breadth and depth of the agent's exploration area, respectively, which effectively enhances data diversity.

\noindent
\textbf{Collection of Image-Action Pair Sequences.}
We use a modern 3D game engine~\citep{ymzx} as our data collection environment, where the exploration agent collects interactive video data consisting of image-action pair sequences $\{o_t, a_t\}$. 
High-precision image-action pair sequences are crucial for the interactive video generation model to learn correct interaction responses. 
%
%
Given the complexity of the environment engine, we align actions and images by recording timestamps. 
Specifically, during the exploration, we capture screenshots at the exact timestamps when the agent performs actions, simultaneously saving the action signals alongside their corresponding high-resolution images. 
This approach ensures high-precision $\{o_t, a_t\}$. 
%

\noindent
\textbf{Data Filters.}
Due to limitations in data collection equipment, some data may experience lag or rendering failures. 
Additionally, as the camera view changes, parts of the data may be obscured by scene elements, making large areas of the image invisible. 
We also encounter data that does not conform to game rules (i.e., interactive mechanics). 
These noisy data will negatively impact the model's learning process. 
To remove low-quality data from the vast amount of raw interactive video data, we apply a series of three filters: the visual filter, the anomaly filter, and the rule filter. 
Below is a detailed introduction to these filters:

The \textit{visual filter} is responsible for removing image data that exists rendering failures or is occluded by scene elements. 
Such image data typically exhibits low color variance. 
Therefore, we compute the mean color variance of all image data in a segment and compare it to a threshold. 
Specifically, let the color variance of each frame be $\sigma_i^2$. We calculate the mean as follows:
$
\bar{\sigma}^2 = \frac{1}{n} \sum_{i=1}^{n} \sigma_i^2.
$
If $\bar{\sigma}^2$ is less than a certain threshold $\tau_1$, the data is deemed to be a rendering failure or occluded and is discarded. 
The \textit{anomaly filter} is designed to identify data that indicates video stuttering. This type of anomalous data often includes an excess of redundant frames, resulting in noticeable stuttering and an abnormally high frame count. 
Therefore, the anomaly filter removes samples with frame counts exceeding a threshold. 
Let the frame count of a video segment be $f$. If $f > \tau_2$, the segment is considered anomalous and filtered out. 
This process prevents the model from learning abnormal video behavior from stuttering data. 
The \textit{rule filter} aims to remove data with inconsistent environmental interaction rules. 
Such inconsistencies are primarily due to the characteristics of the game engine. 
For example, during the "preparation phase", mechanisms in the scene are inactive, meaning non-interactive, leading to inconsistencies in the data regarding mechanism properties. 
After processing through these three filters, we obtain a set of interactive video data with highly consistent mechanics.

\noindent
\textbf{Data Balancer.}
Strongly biased data can cause the model to overfit to specific scenarios, hindering its ability to learn general and diverse interactive video generation. 
After filtering the data, we perform balanced sampling to ensure a relatively even distribution across various dimensions. 
Specifically, when recording image-action pairs, we also capture additional information from the engine at timestep $t$ (such as xyz coordinates, whether the agent is alive, whether it is colliding, etc.) and store it alongside the image-action pairs. 
Following the PlayGen~\citep{playgen}, we perform the balanced sampling method across these attributes to ensure that the sampled subset possesses balanced transition characteristics overall, such as a uniform positional distribution.

\subsection{High-Quality Interactive Video Data}

\noindent
\textbf{1080P High-Resolution Images.}
When launching the 3D game engine, we set the resolution to 1920$\times$1080, allowing us to capture and save high-resolution 1080P images during exploration. 
Consequently, this places higher demands on the performance of the graphics card. We use an NVIDIA RTX 4060 graphics card for rendering the images within one engine.

\noindent
\textbf{30FPS High-Frame-Rate Videos.}
When an agent performs actions at a high frame rate, the game engine may not respond quickly enough, causing inconsistencies in the image-action pair sequences. 
To address this and collect high-frame-rate interactive video data, we use a simple yet effective strategy: action interpolation. 
The exploration agent sends actions 10 times per second and records their timestamps. 
Simultaneously, we capture screenshots at 30 times per second and match them to the corresponding actions using these timestamps. 
Consequently, only 10 frames have corresponding actions, which we refer to as valid frames. 
For the two frames between each pair of adjacent valid frames, we assign the action value of the preceding valid frame.
This approach allows us to generate interactive video data at 30 frames per second (FPS).

\noindent
\textbf{High-Precision Image-Action Pair Sequences.}
By aligning image and action sequences through matching timestamps, the precision of the collected interactive data is significantly enhanced. 
For instance, an action recorded at frame $t$ is captured in the subsequent 1-2 image frames. 
These nearly delay-free image-action pair sequences enable models to learn accurate responses to actions effectively.

\noindent
\textbf{Diverse Action Space.}
Beyond basic actions like moving up, down, left, right, and jumping, we integrate skills such as swooping and view-turning actions like left and right rotations. 
This expansion enriches the action space and enhances user freedom. 
Additionally, compared to a static viewpoint, view-turning actions enable the exploration agent to discover more scenes, resulting in a broader range of image data. 
%

\noindent
\textbf{Diverse Interactive Scenarios.}
We use the 3D game environment as our testbed because it not only allows for high-precision image-action pair sequences but also offers over 90 different styles of scenarios for the exploration agent to gather interactive video data. 
These scenarios cover grasslands, castles, rainforests, valleys, and more. Notably, the out-of-domain test scenarios mentioned later do not include the in-domain scenarios mentioned above.


\begin{table}[htbp]
\centering
\caption{Comparison with other interactive datasets.}
\label{tab:dataset_comparison}
\begin{tabular}{@{} l *{5}{c} @{}}  
\toprule
& \textbf{Resolution} & \textbf{FPS} & \textbf{Frame-wise} & \textbf{Action Space}  & \textbf{Scale}\\
\midrule
The Matrix \citep{the_matrix}      & 720P          & 60    & $\checkmark$   & 5          & 792M     \\
PlayGen \citep{playgen}            & 128P          & 30    & $\checkmark$   & 5          & 250M  \\
GameGenX \citep{gamegen_x}         & 720P-4K       & 1-24  & $\times$       & $\times$   & 192M       \\
GameFactory \citep{gamefactory}    & 360P          & 16    & $\checkmark$   & 9          & 4M     \\
Matrix-Game \citep{matrix_game}    & 720P          & 16    & $\checkmark$   & 7          & 50M       \\
Ours                               & 1080P         & 30    & $\checkmark$   & 8          & 400M \\
\bottomrule
\end{tabular}
\end{table}

\subsection{Data Summary}
Utilizing the automated data collection pipeline constructed above, we collect a large-scale, high-quality set of interactive video data. 
Specifically, we collect over 400 million frames of interactive video data, covering more than 90 different styles of scenarios. 
This dataset includes high-quality sequences of image-action pairs and a diverse action space, ensuring correct action responses and high interactivity. 
We compare our collected dataset with other interactive datasets, as shown in Tab.~\ref{tab:dataset_comparison}.
Overall, the data we collected has the following core features: high resolution (1080P), high frame rate (30FPS), high precision (alignment of images and actions), and diversity (in both action space and scenarios).

\section{Methods}
\subsection{Yan-Sim: AAA-level Simulation}
\label{sec:sim}

\subsubsection{Model Architecture}
\texttt{Yan-Sim} leverages Stable Diffusion (SD) \citep{rombach2022high} as its foundational architecture. We achieve substantial improvements in modeling efficiency for dynamic in-game interactions through three key modifications: increasing the Variational Autoencoder (VAE) \citep{kingma2013auto} compression ratio, adapting the diffusion process for real-time interactive inference, and implementing lightweight structural modifications and inference-time optimizations. These innovations enable \texttt{Yan-Sim} to simulate the world at 1080P resolution, achieving speeds of 50–60 FPS.

\textbf{VAE Design} The VAE encodes raw input images into a latent representation, enabling efficient diffusion model training and inference. To increase inference speed, we enhance spatial compression by augmenting the VAE encoder with two single-layer down blocks, raising the spatial downsampling factor from 8 to 32.
Consecutive video frames are concatenated along the channel dimension to achieve a temporal downsampling factor of 2. As a result, we increase the compression rate of VAE from $1 \times 8 \times 8$ to $2 \times 32 \times 32$.
The increased spatial and temporal compression necessitates higher information density per latent token, so we expand the latent channel dimension $C$ to 16. 
Since only the decoder's model latency is considered during inference, lightweight decoders are essential. Thus, we prune each up block by reducing one layer, while also adding a single-layer up block and a pixel-shuffle layer \citep{shi2016real} to maintain alignment with the encoder's downscaling factor. These adjustments achieve an optimal balance between visual quality and inference speed.

\begin{figure}[t]
\centering
\includegraphics[width=0.9\textwidth]{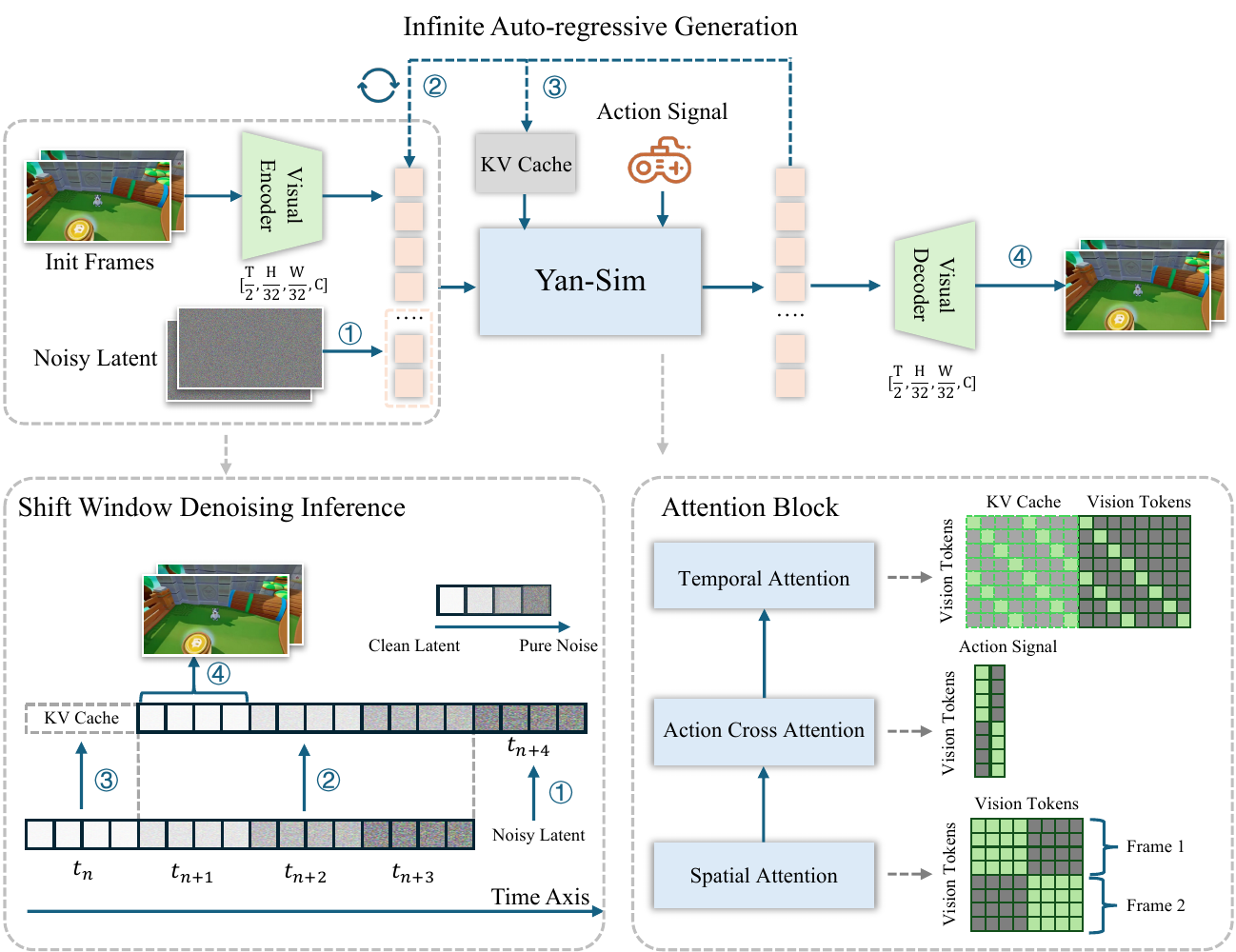}
\caption{
The overall framework of \texttt{Yan-Sim}. The top part of the image depicts \texttt{Yan-Sim}'s entire inference flow, where init frames are only provided during initial inference.
}
\label{fig:yan_sim_framework}
\end{figure}

\textbf{Diffusion Model Design} We adapt the diffusion model for autoregressive, frame-by-frame inference. As shown in Fig. \ref{fig:yan_sim_framework}, the diffusion model includes three types of attention blocks. Spatial attention models dependencies between tokens at different spatial positions within the same frame.
The model takes both image latents and frame-level action signals as input. Action signals are processed by a multi-layer perceptron (MLP) to generate a 768-dimensional token for each frame, which is injected with each token exclusively attending to its corresponding visual frame via action cross-attention.
Following prior work \citep{zhou2022magicvideo, guo2023animatediff, wang2023modelscope}, we introduce 1D temporal attention to model inter-frame dependencies. To enable autoregressive generation, this temporal attention is causal: a token from frame $F_t$ can only attend to tokens from previous frames $F_{<=t}$.
Unlike bidirectional diffusion models, which are constrained to fixed-length sequences, our causal framework supports iterative frame-by-frame prediction. This capability is essential for achieving real-time interactivity and low-latency generation.

\subsubsection{Training}
The training process involves two sequential stages: VAE training and diffusion model training. Due to the abundant data resources and explicit world mechanics of the game environment, we choose the modern 3D game \citep{ymzx} as our world simulation testbed.


\textbf{VAE Training} Given two consecutive frames concatenated along the channel dimension $x \in \mathbb{R}^{H\times W\times 6}$, where $H$ is 1024 and $W$ is 1792 for our 1080P implementation, the encoder encodes $x$ into a latent $z \in \mathbb{R}^{h\times w\times 16}$. The model is optimized using a combined loss function consisting of Mean Squared Error (MSE) and Learned Perceptual Image Patch Similarity (LPIPS) \citep{zhang2018unreasonable}.

\textbf{Diffusion Model Training} Following the SD framework \citep{stable_diffusion}, the diffusion model is trained using the Denoising Diffusion Probabilistic Models (DDPM) paradigm \citep{ho2020denoising}. To achieve autoregressive inference, we employ the Diffusion Forcing strategy \citep{chen2024diffusion, song2025history}: during training, independent noise is added to the latent representation of each frame in a sampled video clip. The first frame, lacking prior context, is treated as a clean ground-truth conditioning signal and receives no noise. To mirror the inference process where noise levels progressively increase along the time axis, we implement a staged noise addition schedule during training.

\subsubsection{Inference}
The primary objectives during inference are minimizing latency and maximizing throughput. To achieve this, we use a DDIM sampler \citep{song2020denoising} that reduces denoising steps to 4, ensuring minimal iterations while maintaining satisfactory visual quality. 
Fig. \ref{fig:yan_sim_framework} shows the inference loop: Given an initial frame, the VAE compresses it into the latent space. A user-provided action signal serves as the conditioning input. The initial latent is then concatenated with a noisy latent and passed into the diffusion model. The diffusion model denoises the noisy latent, resulting in a clean latent representing the next frame corresponding to the specified action. This clean latent is then decoded by the VAE to produce the next frame, which becomes the input latent for the subsequent inference step. This continuous loop allows for real-time, interactive content generation. To further optimize performance and reduce latency, several strategies are implemented, including shift window denoising inference, pruning \& quantization, and additional optimizations. These improvements enhance model efficiency, enabling faster and more responsive video generation.

\textbf{Shift Window Denoising Inference}  
Traditional denoising methods require multiple iterations per sample, causing latency to increase linearly with the number of denoising steps. To reduce latency, we propose a shift window denoising inference pipeline, where each inference step processes a window of frames. Within this window, frames vary in noise levels—earlier frames are cleaner, while later frames are noisier. At each step, a pure noise latent is concatenated to the input latent(as shown in Fig. \ref{fig:yan_sim_framework}\textcircled{1}). Each inference step produces a clean latent, which is then decoded back into RGB images (Fig. \ref{fig:yan_sim_framework}\textcircled{4}). Additionally, we implement KV caching to store historical states and avoid redundant computations (Fig. \ref{fig:yan_sim_framework}\textcircled{3}). This method effectively reduces the average latency per generated frame.

\textbf{Pruning \& Quantization} To meet the real-time requirements of AAA-simulating, we apply structural pruning to the UNet architecture and quantize both weights and activations to FP8 across all GEMM operations. This FP8 quantization delivers a $1.5\times - 2\times$ speedup over FP16, contributing to a $1.18\times$ overall inference speedup with minimal loss in generation quality. Additionally, capturing the entire inference pipeline as a single CUDA graph eliminates kernel-launch overhead, while Triton-based custom kernels optimize thread-block geometry and memory access. By enabling torch.compile with max-autotune mode, these optimizations are automatically applied, further boosting performance by $1.15\times$.
Additionally, to avoid the latency caused by serial inference, we deploy the two models on separate GPUs and use a queue to transfer the diffusion-generated latent to the VAE.

\subsubsection{Evaluation}
In this subsection, we qualitatively evaluate the AAA-level simulation performance of \texttt{Yan-Sim} across four dimensions: visual quality, motion consistency, world physics, and long video generation capability. Additionally, we compare it with other game world simulation technologies.

\begin{figure}[t]
\centering
\includegraphics[width=0.99\textwidth]{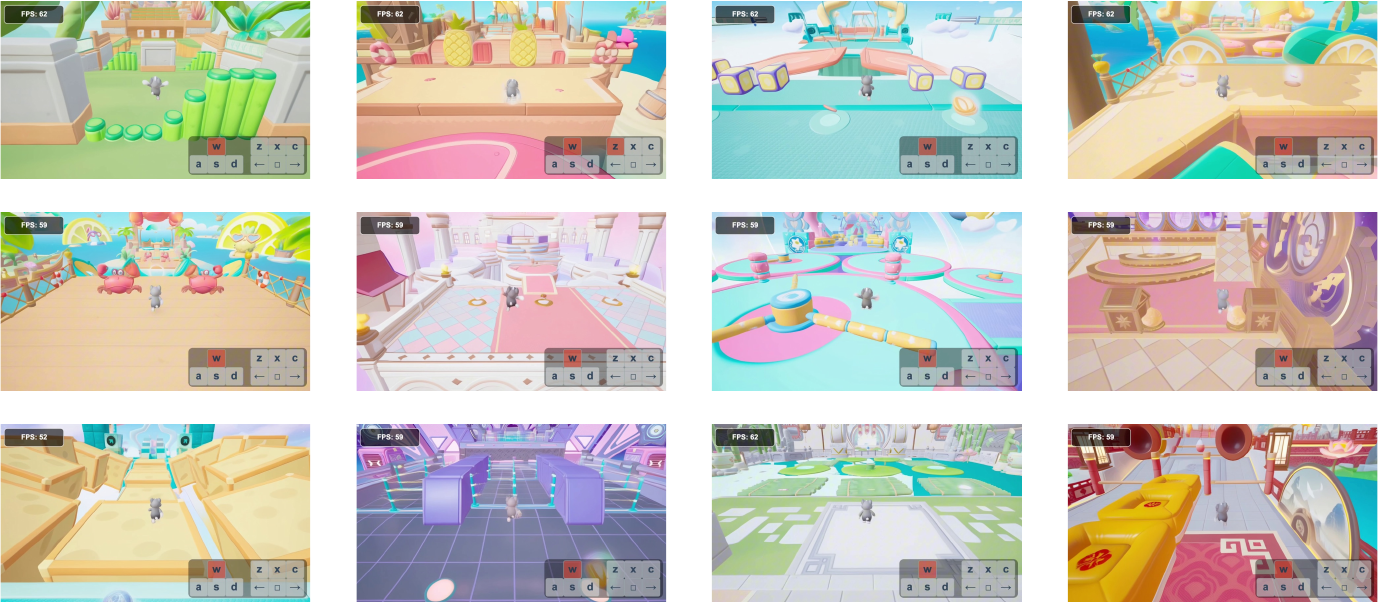}
\caption{
The visualization results of visual quality using \texttt{Yan-Sim}. \texttt{Yan-Sim} possesses the ability to simulate diverse styles and interactive scenarios.
}
\label{fig:Yan_Sim_Evaluation_visual_quality}
\end{figure}

\textbf{Visual Quality} The simulation faces two major challenges in terms of visual quality. First, the game features a wide range of artistic styles, and the simulation must faithfully reproduce all of these diverse visuals. Second, it is required to achieve this reproduction at high resolution. As shown in Fig. \ref{fig:Yan_Sim_Evaluation_visual_quality}, \texttt{Yan-Sim} successfully renders the intricate details of eight different levels in high resolution and high FPS, accurately reflecting the vibrant scenes.

\begin{figure}[t]
\centering
\includegraphics[width=0.99\textwidth]{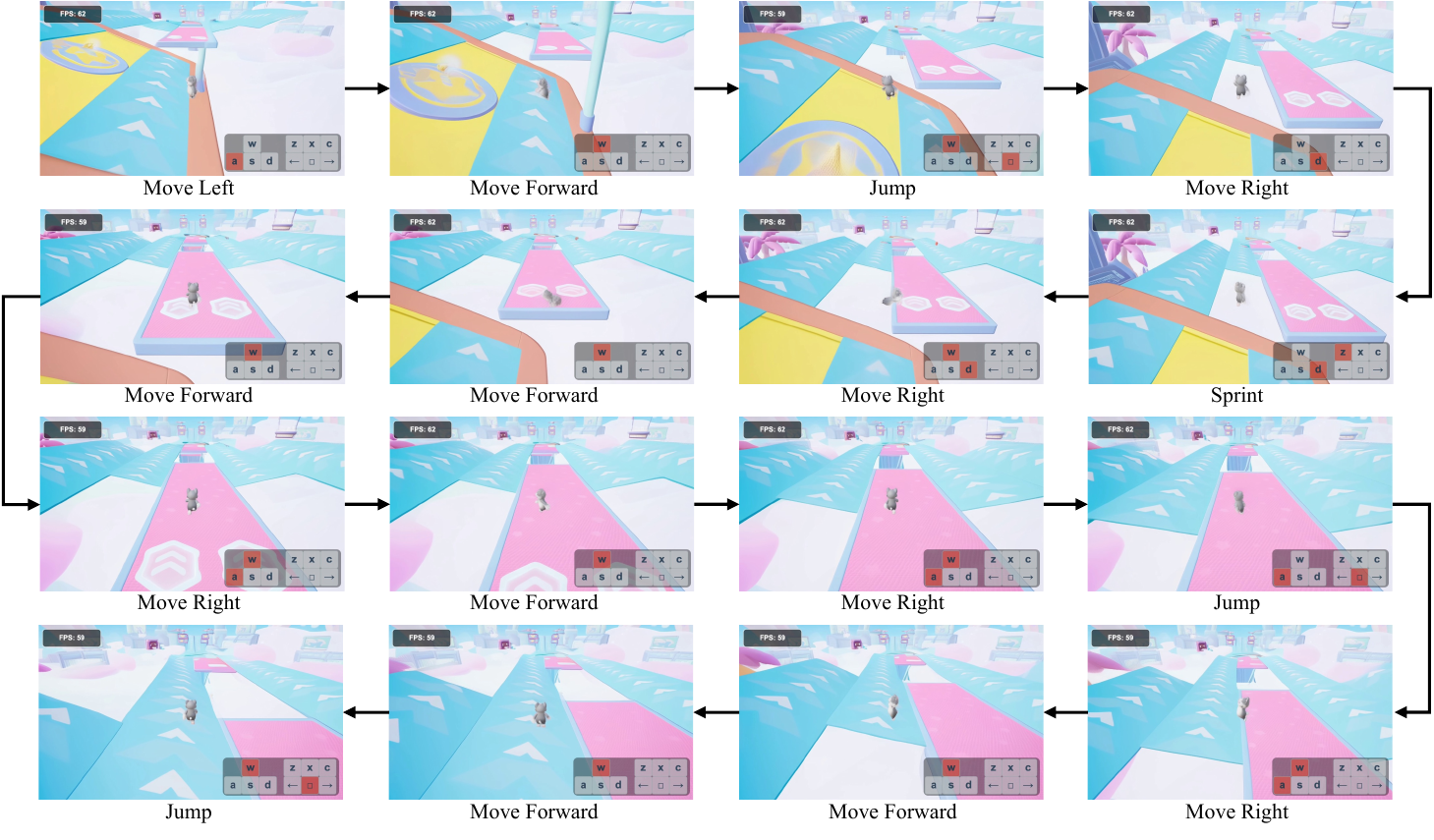}
\caption{
The visualization results of motion consistency using \texttt{Yan-Sim}. The top-left corner marks the starting frame. Actions performed in each frame will be reflected in the subsequent frame. Due to the high fps, differences between consecutive frames are minor. For clarity, some frames have been skipped in this demonstration.
}
\label{fig:Yan_Sim_Evaluation_motion_consistence}
\end{figure}

\textbf{Motion Consistency} Interactive simulation emphasizes the ability for users to control the generated visuals with frame-level action signals. As shown in Fig. \ref{fig:Yan_Sim_Evaluation_motion_consistence}, \texttt{Yan-Sim} responds accurately to different user inputs, generating the correct feedback in the subsequent frames. Thanks to high FPS and low latency, the experience closely mirrors real gameplay.

\begin{figure}[t]
\centering
\includegraphics[width=0.99\textwidth]{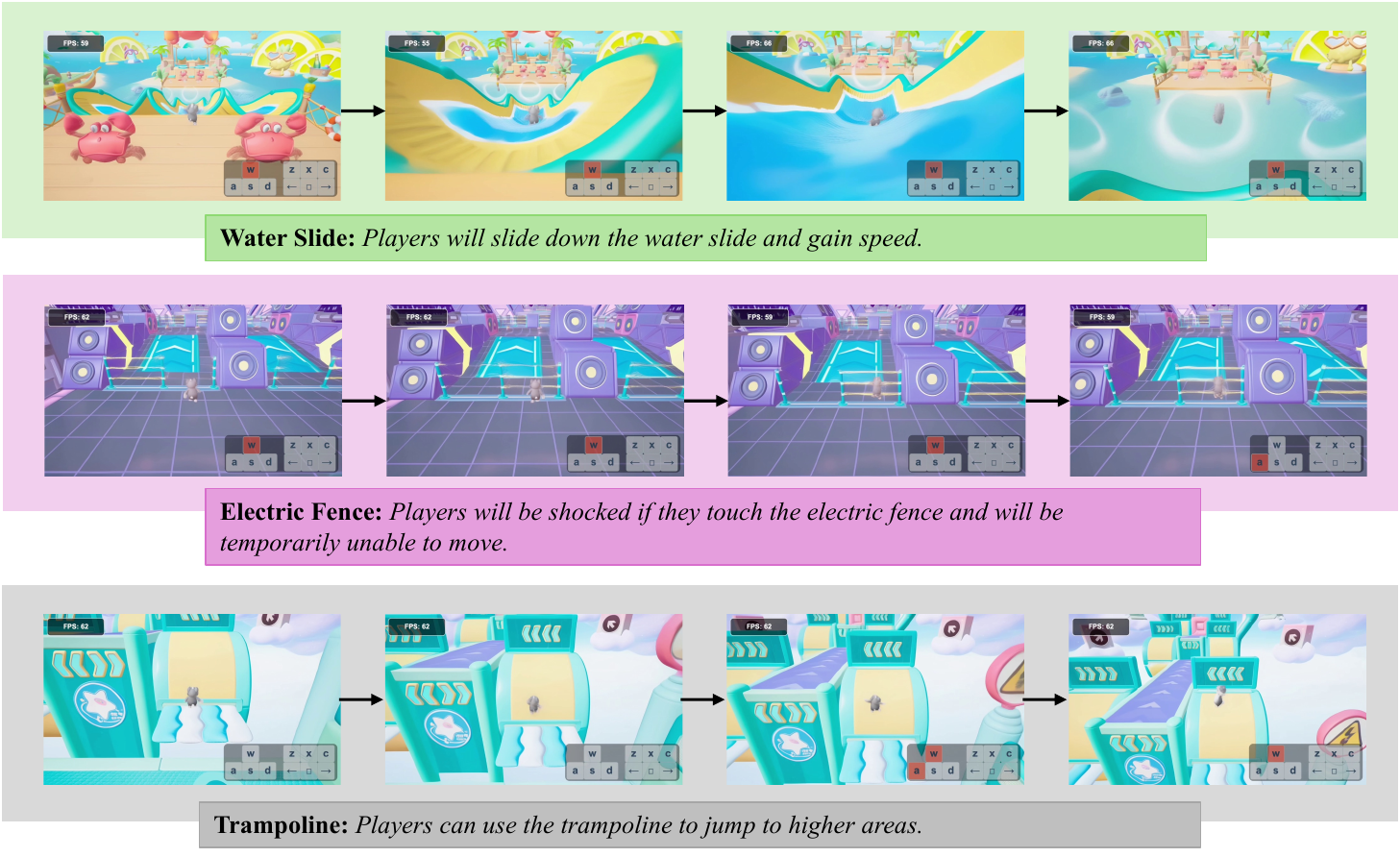}
\caption{
The visualization results of accurate mechanism simulation using \texttt{Yan-Sim}. 
}
\label{fig:Yan_Sim_Evaluation_world_phyisc}
\end{figure}

\textbf{Accurate Mechanism Simulation} The simulation also aims to realistically reproduce the physical characteristics. Fig. \ref{fig:Yan_Sim_Evaluation_world_phyisc} demonstrates various interactive effects, for example,  \textbf{Water Slide}: the character’s natural inertia when sliding down a ramp; \textbf{Electric Fence}: the character being unable to move after touching an electric fence; \textbf{Trampoline}: players can jump higher on the trampoline. These examples showcase \texttt{Yan-Sim}’s accurate replication of the world’s physics.

\textbf{Long Video Generation Capability} \texttt{Yan-Sim} can generate long interactive videos. The visual consistency remains high throughout, further validating \texttt{Yan-Sim}’s ability to produce continuous high-quality content.

Additionally, in Table \ref{tab:framework-comparison}, we compare \texttt{Yan-Sim} with other interactive simulation systems. \texttt{Yan-Sim} stands out due to its high resolution, low latency, and high FPS—core attributes that provide a solid foundation for achieving AAA-level simulation quality.


\begin{table}[htbp]
\centering
\caption{Comparison with other world simulation works.}
\label{tab:framework-comparison}
\begin{tabular}{@{} l *{4}{c} @{}}  
\toprule
  & \textbf{Video Length} & \textbf{Resolution} & \textbf{Real Time} & \textbf{Low Latency} \\
\midrule
The Matrix \citep{the_matrix}     & Infinite          & 720p  & $\checkmark$ (16fps)  & $\times$          \\
PlayGen \citep{playgen}      & Infinite          & 128p  & $\checkmark$ (20fps)  & $\checkmark$ (0.05s) \\
Genie 2 \citep{genie2}       & 10 - 20s          & 360p  & $\times$             & $\times$          \\
GameFactory \citep{gamefactory}   & Infinite          & 640p  & $\times$             & $\times$          \\
Matrix-Game \citep{matrix_game}    & Infinite          & 720p  & $\times$             & $\times$          \\
Genie 3 \citep{genie3}       & few minutes  & 720p  & $\checkmark$ (24fps) & $\checkmark$       \\
Yan-Sim        & Infinite          & \textbf{1080p} & $\checkmark$ (\textbf{60fps}) & $\checkmark$ (0.11s) \\
\bottomrule
\end{tabular}
\end{table}
\subsection{Yan-Gen: Multi-Modal Generation}
\label{sec:gen}

\begin{figure}[t]
\centering
\includegraphics[width=0.99\textwidth]{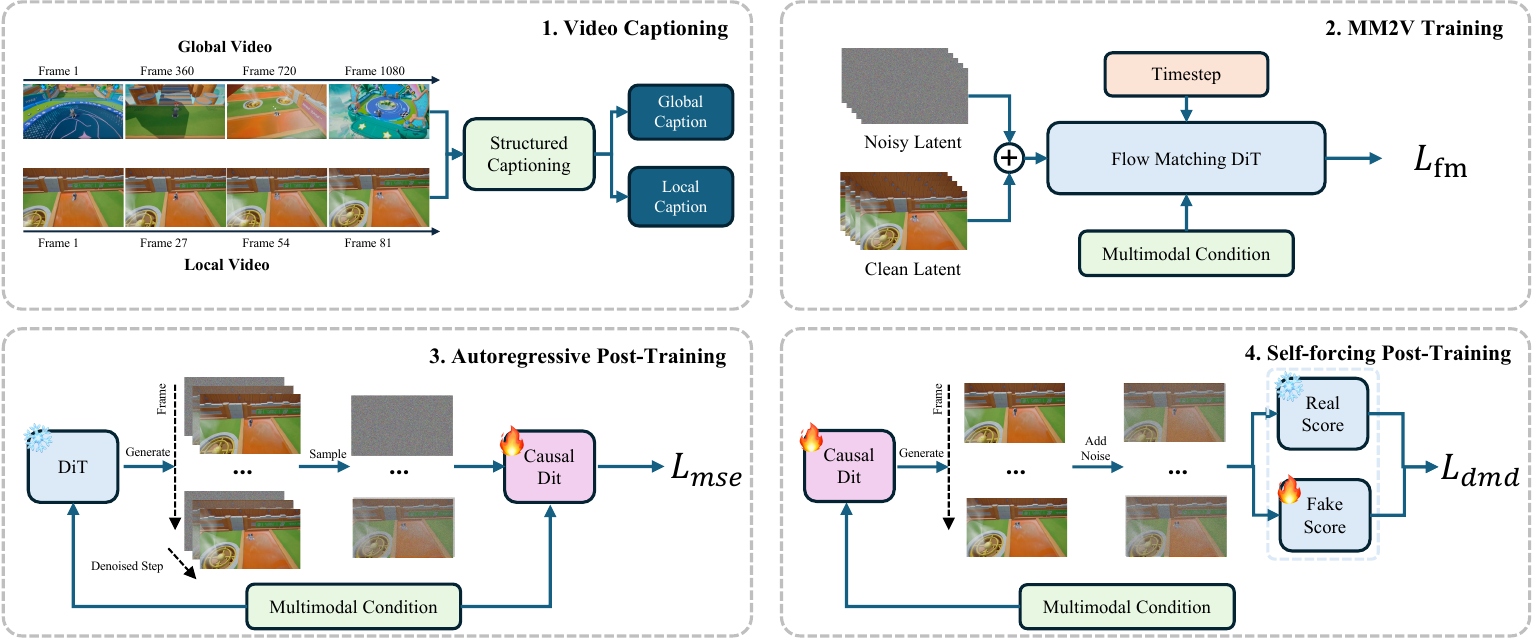}
\caption{
The training pipeline of \texttt{Yan-Gen}, a real-time interactive world generation framework. It includes four stages: hierarchical video captioning, multi-modal conditions to interactive video generation training, auto-regressive post-training for infinite generation, and self-forcing post-training for efficient inference.
}
\label{fig:yan_gen_train}
\end{figure}

In this section, we present \texttt{Yan-Gen}, a novel framework for real-time, interactive world generation. To address the critical challenge of long-term semantic drift inherent in auto-regressive video models, \texttt{Yan-Gen} employs a multi-stage training pipeline, as illustrated in Fig.~\ref{fig:yan_gen_train}. Our process begins by training a foundational world generation model conditioned on a combination of multi-modal inputs: images, text, and frame-wise action signals. This relies on our hierarchical captioning system, which provides both a stable global context to prevent drift and detailed local descriptions for high-fidelity event rendering. Next, to enable continuous, open-ended generation, we convert this base model into a causal, auto-regressive generator by training it on the ordinary differential equation (ODE) trajectories of its predecessor. Finally, to achieve the real-time interaction, we distill this many-step causal model into a highly efficient few-step generator using a self-forcing distillation technique.

\subsubsection{Hierarchical Captioning for World and Local Context Modeling}
\label{sec:yan_gen_hier_caption}
A fundamental challenge in generating long, coherent video sequences is the "anti-drifting" problem. Auto-regressive models, which predict subsequent frames based on previously generated ones, are susceptible to error accumulation and exposure bias. Over time, these compounding errors can cause the generated content to "drift" from the initial context, leading to a degradation of visual quality and a loss of temporal and spatial consistency. This is particularly problematic in game environments, where the world must remain stable and consistent as a character navigates it.

To address this, our methodology establishes a stable "world" through a two-level hierarchical captioning pipeline. This approach decouples the static, global environment from the dynamic, local events occurring within it. By providing the generative model with a constant global reference, we anchor the generation process, preventing long-term semantic drift. Simultaneously, a detailed local caption ensures that fine-grained details are rendered with high fidelity. This dual-context strategy is crucial for producing videos that are both globally coherent over extended durations and locally accurate to fine-grained events.

\paragraph{Global Captioning: Defining the Static World} The first level of the hierarchy creates the static world model. A single global caption is generated for an entire world, based on a video of the environment's traversal. This caption acts as a constant, unchanging point of reference for all subsequent video generation within the world. It aims to capture the world's unchanging characteristics, guaranteeing long-term consistency. The global caption outlines three main elements: the global layout, which details the game world's topology, including its primary regions and interconnections; the visual theme, which covers the overall aesthetic, such as the color palette, main materials, and architectural style; and base lighting and weather, which describes the consistent ambient environmental conditions throughout the world. This static, high-level description provides a stable basis, preventing the model from drifting during continuous auto-regressive inference.

\paragraph{Local Captioning: Grounding Dynamic Events} The second level of the hierarchy provides the dynamic, local context. For each video clip, a local caption is generated to describe the transient events occurring within that specific window. This detailed, time-sensitive information is essential for training the model to accurately align the distribution of text with the distribution of video data, enabling precise rendering of dynamic environmental interactions. The local caption describes the local scene, which includes immediate surroundings and environmental elements visible within the camera's view; the interactive objects, encompassing any objects whose state visibly changes during the clip, capturing real-time environmental interactivity; and critical events like character death or task completion. This local description ensures that the model output is precisely aligned with the fine-grained details of the ongoing interaction.

\paragraph{} In this paper, we adopt Qwen2.5-VL~\citep{bai2025qwen2} to perform hierarchical captioning. A typical environment's traversal video is 1 minute long, and a local video clip length is 3 seconds. In total, we label 98 million frames for training.
\begin{figure}[t]
\centering
\includegraphics[width=0.9\textwidth]{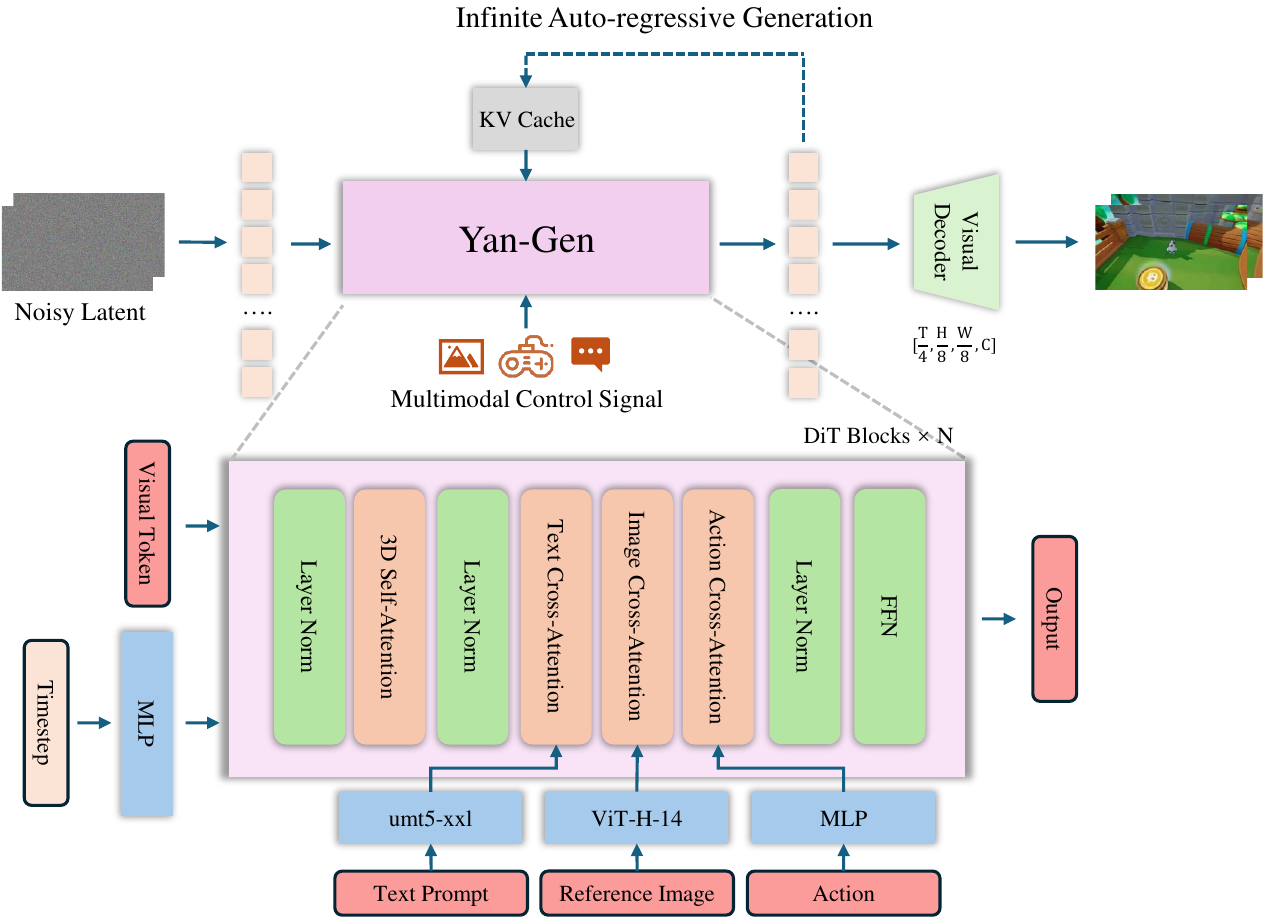}
\caption{
The model architecture of \texttt{Yan-Gen}. The multi-modal conditions are injected via decoupled cross-attention layers.
}
\label{fig:yan_gen_arch}
\end{figure}

\subsubsection{Multi-modal Conditions to Interactive Worlds}
\label{sec:yan_gen_mm2v}
To construct the real-time interactive generation pipeline, \texttt{Yan-Gen} builds upon a pre-trained Wan model~\citep{wan2025wan}, which employs a cross-attention mechanism to embed the input text or image conditions. We first adapt this model to the interactive video data distribution by fine-tuning all its linear layers with Low-Rank Adaptation (LoRA) for both text-to-video and image-to-video generation. During this adaptation phase, we leverage our hierarchical captions to explicitly address the "anti-drifting" problem. The model is conditioned on the concatenated global and local captions, providing the dense supervision necessary for precise text-to-video feature alignment. With a probability of $p=0.1$, the model is provided with only the global caption, which enforces the model to generate a plausible video sequence even with minimal, high-level guidance.

Next, we introduce and train an interactive module to incorporate action-based conditions. While using identical hierarchical captions, action-related descriptions are dropped. To achieve fine-grained, temporally coherent video control, this module features a simple yet effective action encoder. The encoded action sequences are then injected into each block of the DiT via a separate cross-attention layer, guiding the flow matching process. The model detail is shown in Fig.~\ref{fig:yan_gen_arch}. To enforce action causality and precise frame-wise control, we employ an attention mask similar to that in \texttt{Yan-Sim}. Crucially, to ensure the interactive module learns a style-independent control mapping that generalizes effectively to other worlds, all other model parameters remain frozen during this training phase. To further improve robustness and enable classifier-free guidance, we randomly replace action sequences with null actions with a probability of $p=0.1$.

\texttt{Yan-Gen} is trained to predict the velocity following Rectified Flows~\citep{esser2024scaling}, thus the loss function can be formulated as
\begin{equation}\label{rf_loss}
\mathcal{L}_{fm} = \mathbb{E}_{x_0, x_1, c_{txt}, c_{img}, c_{act}, t}||u(x_t, c_{txt}, c_{img}, c_{act}, t; \theta)-v_t||^2,
\end{equation}
where $c_{txt}$ is the umt5-xxl~\citep{chung2023unimax} text embedding sequence of 512 tokens long, $c_{img}$ is the ViT-H-14~\citep{radford2021learning} image embedding sequence of 257 tokens long, $c_{act}$ is the action sequence of video length, $\theta$ is the model weights, and $u(x_t, c_{txt}, c_{img}, c_{act}, t; \theta)$ denotes the output velocity predicted by the model. Upon completion of this training, \texttt{Yan-Gen} can generate fixed-length videos guided by a reference image, descriptive text prompts, and a sequence of user actions via
\begin{equation}
\begin{aligned}
u_{output} &= u(x_t, \varnothing, c_{img}, \varnothing, t; \theta) \\
&\quad + g_{uncond} \cdot \bigl(u(x_t, c_{txt}, c_{img}, c_{act}, t; \theta) - u(x_t, \varnothing, c_{img}, \varnothing, t; \theta)\bigr).
\end{aligned}
\end{equation}

\subsubsection{Auto-regressive Post-training}
\label{sec:yan_gen_arpt}
To enable real-time interaction in \texttt{Yan-Gen}, we follow \citet{yin2025slow} to convert \texttt{Yan-Gen} into a few-step causal model. The process begins with auto-regressive initialization. We collect prompts from the Yuanmeng World dataset and use \texttt{Yan-Gen} trained in Sec .~\ref {sec:yan_gen_mm2v} to perform multi-step denoising via flow matching. This generates a ode trajectory of samples $\{ x_t \}_{t=T}^{0}$, and we select a subset of 4 noise samples that are used in our 4-step generator $\{ x_T, x_{t_1}, x_{t_2}, x_{0} \}$. Here, $x_T$ represents pure noise, and $x_0$ is the clean latent.

Next, we modify \texttt{Yan-Gen} into a causal model by employing block causal attention in the self-attention layer (ensuring each frame's token can only attend to prior frames). Causal \texttt{Yan-Gen} is trained on the 4-step ode trajectory dataset with the diffusion forcing paradigm. During training, each frame’s latent in a clip is randomly sampled from one of 4 predefined noise levels, and the generated output is compared to the clean latent $x_{0}$ using a regression loss, which can be formulated as 
\begin{equation}
\begin{aligned}
\mathcal{L}_{mse} = \mathbb{E}_{x, c_{txt}, c_{img}, c_{act}, t^{i}} \left\| u^\prime\left( \{ x_{t^{i}}^{i} \}_{i=1}^{N}, \{ t^{i} \}_{i=1}^{N} \right) - \{ x_{0}^{i} \}_{i=1}^{N} \right\|^{2},
\end{aligned}
\end{equation}

where $u^\prime$ is the causal \texttt{Yan-Gen}, $N$ is the latent length, and $t^i$ can be selected from $\{ T, t_1, t_2, 0 \}$.

\subsubsection{Self-forcing Post-training}
\label{sec:yan_gen_sfpt}
We train causal \texttt{Yan-Gen} using distribution matching distillation (DMD) \citep{yin2024one} to extract a few-step generator. DMD minimizes the KL divergence between the generator’s output distribution and the true data distribution.

DMD uses two score functions initialized from the \texttt{Yan-Gen} trained in Sec .~\ref {sec:yan_gen_mm2v}. The real score is frozen, representing the true data distribution, while the fake score continuously fits to the generator's distribution by learning from the generator's output. The DMD loss can be formulated as
\begin{equation}
\begin{aligned}
\nabla_{\phi}\mathcal{L}_{dmd} & \triangleq \mathbb{E}_{t}\left( \nabla_{\phi}\mathbf{KL}\left( p_{\text{gen},t} \| p_{\text{data},t} \right) \right) \\
& \approx -\mathbb{E}_{t}\left( \int \left( s_{\text{real}}\left( u(u_{\phi}'(\epsilon), t), t \right) - s_{\text{fake}}\left( u(u_{\phi}'(\epsilon), t), t \right) \right) \frac{du_{\phi}'(\epsilon)}{d\phi}  d\epsilon \right),
\end{aligned}
\end{equation}

where $\phi$ represents the learnable parameter of causal \texttt{Yan-Gen}, and $\epsilon$ is random Gaussian noise. DMD2 \citep{yin2024improved} converts the one-step generator into a few-step generator by replacing pure noise with noisy sample data. Self-forcing \citep{huang2025self} trains the generator using an auto-regressive paradigm to generate training samples, reducing the cumulative errors introduced by teacher forcing.

The distilled few-step \texttt{Yan-Gen} enables real-time interaction at 12-17 FPS on a single NVIDIA H20. With unified sequence parallel inference, \texttt{Yan-Gen} can scale to 30 FPS using 4 NVIDIA H20. After post-training, \texttt{Yan-Gen} can generate infinite content and respond to action signals in real-time, or adapt the content by switching prompts.

\subsubsection{Evaluation}
In this section, we qualitatively evaluate the multi-modal generation ability of \texttt{Yan-Gen} by illustrating text-to-world and image-to-world results. All the results are real-time interactive gameplay by humans, with precise frame-by-frame responsiveness to user actions.

\begin{figure}[t]
\centering
\includegraphics[width=0.99\textwidth]{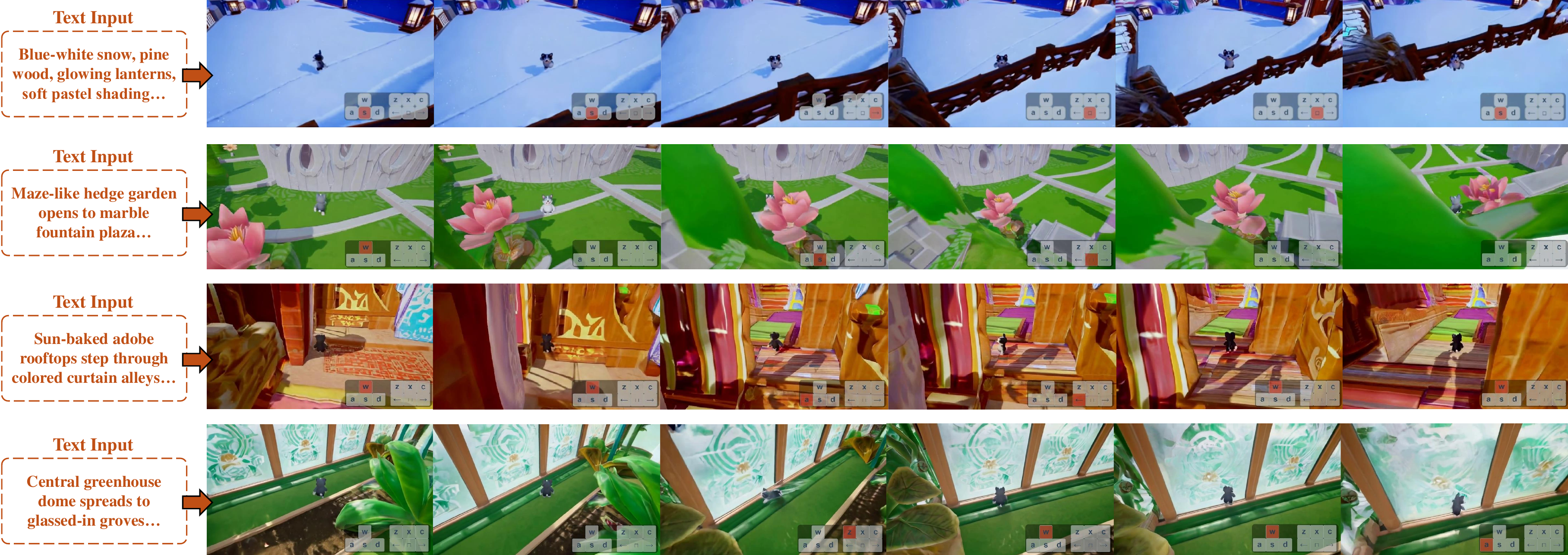}
\caption{
The visualization results of text-to-interaction using \texttt{Yan-Gen}, which demonstrates exceptional real-time physical-aware interaction and instruction following ability.
}
\label{fig:t2v_results}
\end{figure}

\begin{figure}[t]
\centering
\includegraphics[width=0.99\textwidth]{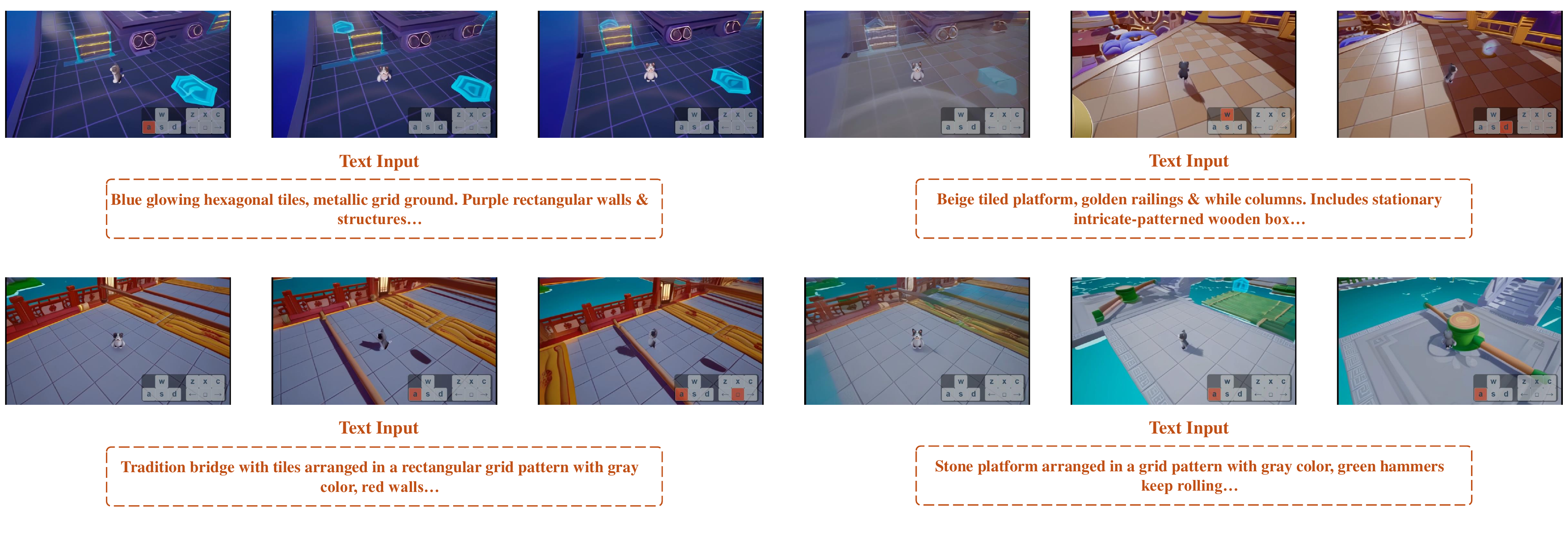}
\caption{
The visualization results of text-guided expansion using \texttt{Yan-Gen}, where \texttt{Yan-Gen} can expand the original scene following the modified scene descriptions in a continuous manner.
}
\label{fig:text_expansion}
\end{figure}

\begin{figure}[t]
\centering
\includegraphics[width=0.99\textwidth]{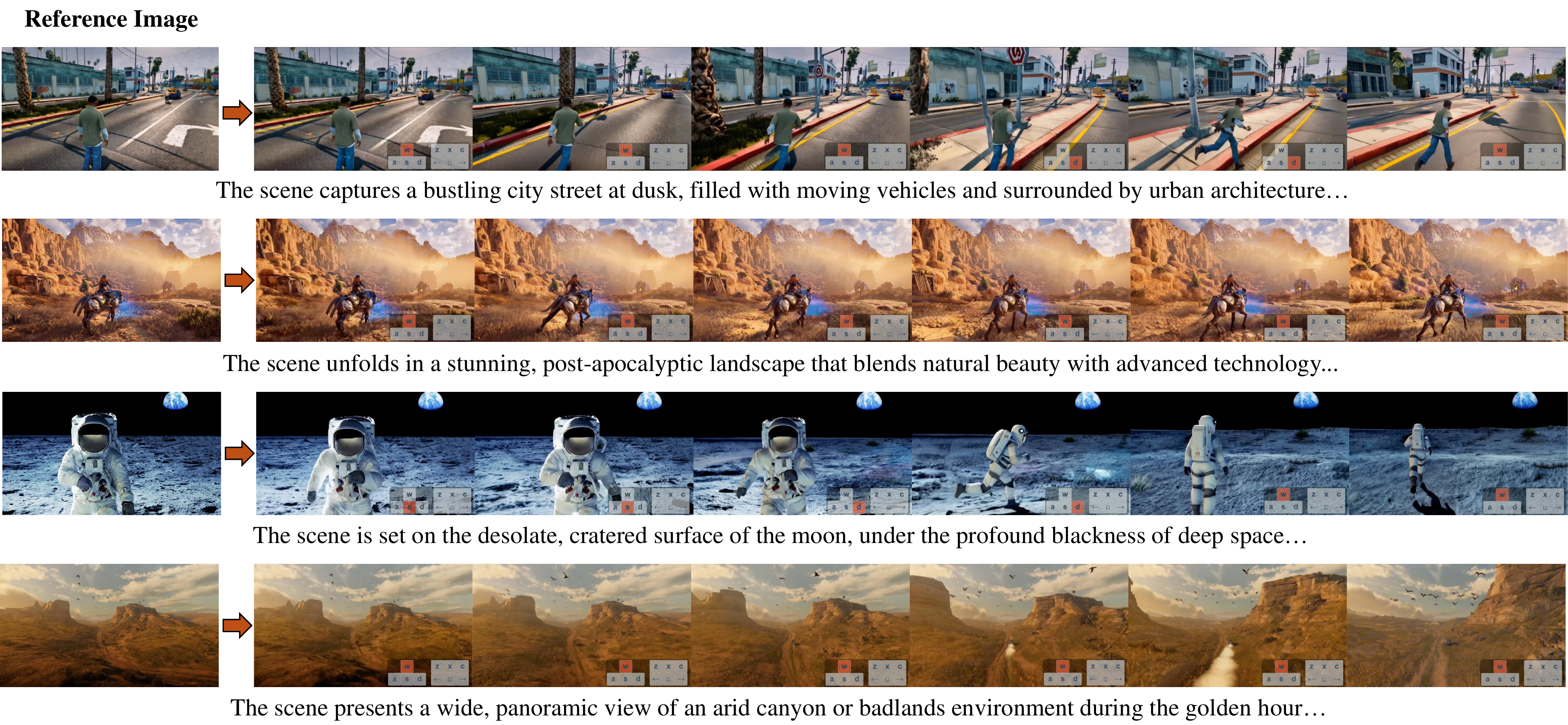}
\caption{
The visualization results of image-to-interaction. Given an initial reference image, \texttt{Yan-Gen} can perform interactive generation based on user-input actions without losing the features of the reference image.
}
\label{fig:i2v_results}
\end{figure}

\begin{figure}[t]
\centering
\includegraphics[width=0.99\textwidth]{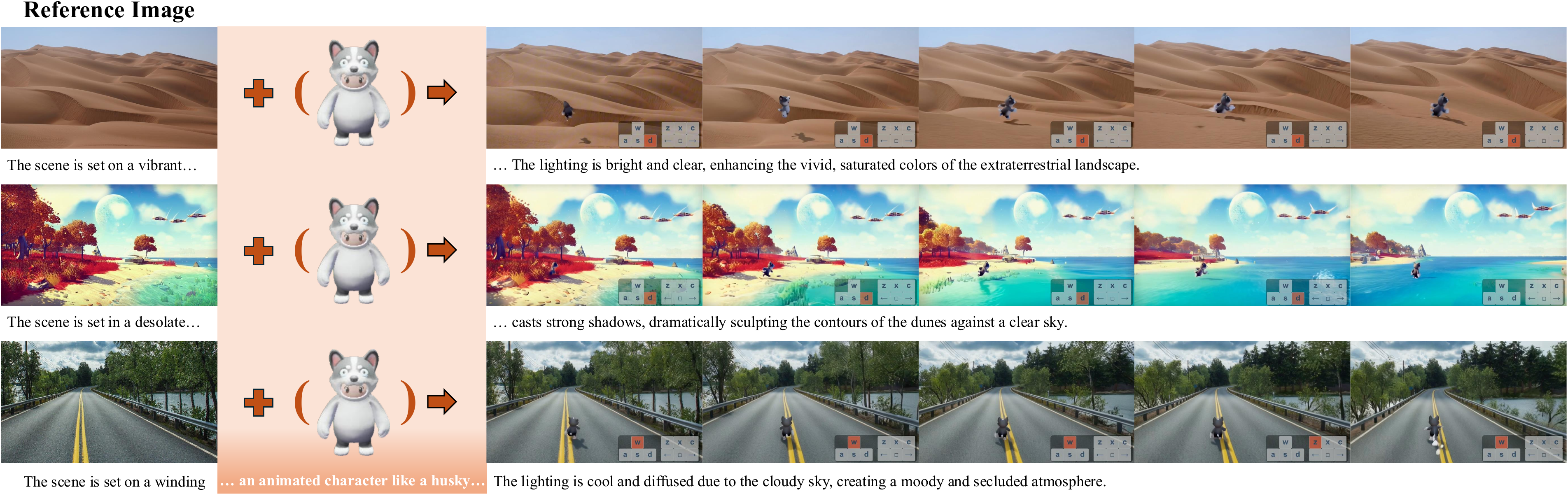}
\caption{
The visualization results of cross-domain fusion. \texttt{Yan-Gen} can fuse out-of-domain images with in-domain subjects, demonstrating its interactive generation capability of cross-domain knowledge fusion.
}
\label{fig:i2v_cross_domain}
\end{figure}

\paragraph{Text-to-interaction.}
As shown in Fig.~\ref{fig:t2v_results}, \texttt{Yan-Gen} generates interactive scenes where a character sprints across snowy terrain and leaps over a railing with realistic momentum; navigates a hedge maze, maintaining spatial awareness even after prolonged occlusion behind foliage before emerging at a marble plaza; dynamically interacts with billowing curtains on sunlit adobe rooftops, adjusting movements to avoid entanglement; and halts upon colliding with glass panes in a greenhouse, distinguishing transparent barriers from foliage. These behaviors reflect the model’s ability to synthesize real-time physics-aligned actions, contextual spatial reasoning, material-aware interaction synthesis, and adaptive obstacle negotiation within immersive text-driven environments.

\paragraph{Text-Guided Expansion.} 
As illustrated in Fig.~\ref{fig:text_expansion}, \texttt{Yan-Gen} implements text-guided scene expansion. During interactive sessions, \texttt{Yan-Gen} dynamically expands the scene when provided with modified textual descriptors. For example, \texttt{Yan-Gen} expands the original disco-themed environment to a classical musician scene. Crucially, this expansion process exhibits continuity, where shared structural elements (e.g., tiling patterns and platform colors) serve as transitional elements that mediate cross-domain scene morphogenesis.

\paragraph{Image-to-interaction.}
As illustrated in Fig.~\ref{fig:i2v_results}, \texttt{Yan-Gen} seamlessly responds to diverse open-world scenarios. These results confirm consistent responsiveness across various scenarios, including urban chaos and technological wilderness, validating the action cross-attention mechanism’s scalability for open-world real-time interactions without domain-specific tuning.

\paragraph{Corss-Domain Fusion.}
As illustrated in Fig.~\ref{fig:i2v_cross_domain}, \texttt{Yan-Gen} is able to perform cross-domain fusion generation. We use out-of-domain reference images and text containing in-domain character descriptions as prompts to guide the model in combining knowledge from different domains. Ultimately, \texttt{Yan-Gen} achieves cross-domain fusion interactive video generation.

\subsection{Yan-Edit: Multi-Granularity Editing}
\label{sec:edit}



In this section, we propose \texttt{Yan-Edit}, a text-driven interactive video editing model that enables users to modify the video content through texts at any moment during interaction. Besides, \texttt{Yan-Edit} supports multi-granularity video content editing, encompassing both structure editing (e.g., adding interactive objects) and style editing (e.g., altering an object's color and texture). In the following, we first introduce the model design and architecture of \texttt{Yan-Edit} in Sec .~\ref {yan_edit:model_archi}. We then describe the training and inference strategy in Sec .~\ref {yan_edit:train_and_infer}. Finally, we present qualitative results to demonstrate the versatile editing capabilities of \texttt{Yan-Edit} in Sec.~\ref{yan_edit:eval}.

\subsubsection{Model Design and Architecture}
\label{yan_edit:model_archi}

\begin{figure}[t]
\centering
\includegraphics[width=0.99\textwidth]{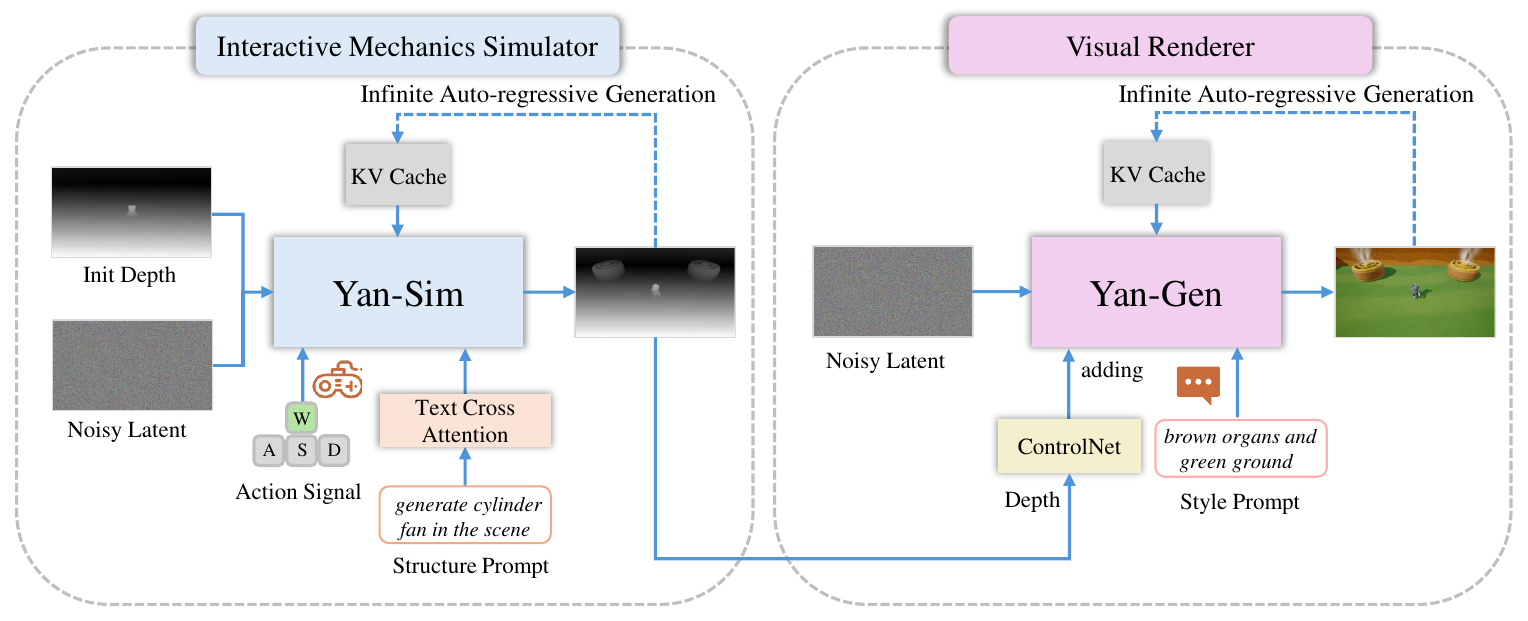}
\caption{
The overall framework of \texttt{Yan-Edit}. It consists of an interactive mechanics simulator and a visual renderer. The interactive mechanics simulator is built upon \texttt{Yan-Sim}, which learns the structure-dependent interactive mechanics based on the depth maps. The visual renderer based on \texttt{Yan-Gen} is responsible for rendering the depth maps generated by the interactive mechanics simulator in versatile open-domain styles. Two types of text prompts are introduced to enable multi-granularity editing: structure prompt for structure editing (e.g., adding interactive objects) and style prompt for style editing (e.g., altering an object's color and texture). 
}
\label{fig:yan_edit_framework}
\end{figure}

Compared to previous video editing~\citep{vace,pix2video,video_p2p} that does not involve real-time user interaction, the main challenge of interactive video editing is how to ensure that the editing content meets the interactivity requirements. For example, in a racing game video, if we change the color of the controlled car from blue to green, how can we maintain the original interactive mechanics of the blue car when the user interacts with the new green car. To realize interactive video editing with accurate interactive mechanics simulation, we propose a hybrid model \texttt{Yan-Edit} that disentangles interactive mechanics simulation from visual rendering.

The overall framework of \texttt{Yan-Edit} is shown in Fig.~\ref{fig:yan_edit_framework}. Specifically, we first extract the depth map for each frame in the interactive video data, which retains the 3D structure of objects in the video and removes its visual rendering information (e.g., the color and texture of objects). Then, we utilize an interactive mechanics simulator to learn the simulation of interactive mechanics based on the depth maps. In this way, we can learn a more general simulation of interactive mechanics that depends on the structures of objects rather than their colors and textures. Next, we employ a visual renderer to render the depth map as the final output video frame. To enable user-friendly editing, we introduce two types of text prompts: a structure prompt for structure editing and a style prompt for style editing. The model architecture of the interactive mechanics simulator and visual renderer is described below.

\paragraph{Interactive Mechanics Simulator.} 
The interactive mechanics simulator needs to learn structure-dependent interactive mechanics based on the depth maps, so we adopt \texttt{Yan-Sim} as the base model of the interactive mechanics simulator, leveraging its high-fidelity simulation of interactive mechanics. Given the initial depth, it generates consecutive depth frames auto-regressively conditioned on the user's per-frame action signal (e.g., keyboard control signal) and structure text prompt. The structure text prompt is integrated into \texttt{Yan-Sim} by inserting a text cross-attention layer after the action cross-attention layer of each attention block.

\paragraph{Visual Renderer.} 
The visual renderer aims to render the depth frames generated by the interactive mechanics simulator. We use style text prompts to control the rendering style (i.e., the color and texture of objects in depth frames). To enable versatile rendering, the visual renderer takes advantage of \texttt{Yan-Gen} for its powerful generation capabilities on various open-domain visual content. The depth frames are injected into \texttt{Yan-Gen} via a ControlNet~\citep{controlnet}. Similar to the structure of VACE~\citep{vace}, the ControlNet consists of several DiT blocks copied from the original DiT blocks in \texttt{Yan-Gen}, and the output of each DiT block in ControlNet is added back to the original DiT blocks in \texttt{Yan-Gen}.

\subsubsection{Training and Inference}
\label{yan_edit:train_and_infer}

We first train the interactive mechanics simulator to learn high-fidelity interactive mechanics simulation on the depth maps. We incorporate structure prompts that describe the scene structure to train \texttt{Yan-Sim}. For precise structural characterization, we segment the interactive video data collected in Sec .~\ref {sec:data} into 81-frame temporal chunks and employ Qwen2.5-VL~\citep{bai2025qwen25vltechnicalreport} to perform three-dimensional structural decomposition, categorizing elements as: (1) dynamic objects, (2) wall structures, and (3) environmental platforms. Following the training paradigm of \texttt{Yan-Sim}, our training protocol comprises two distinct phases. In the first phase, we utilize DepthFM~\citep{depthfm} to extract the depth maps from videos and train the VAE specifically for depth maps. In the second phase, we train the action and text cross-attention layers simultaneously for accelerated convergence, and then freeze the text attention weights while fine-tuning the action ones alone to boost consistency and fidelity. 

Then, we train the visual renderer to learn versatile-style rendering on depth frames generated by the interactive mechanics simulator. Following the training paradigm of \texttt{Yan-Gen}, we should first train a non-casual video generation model conditioned on the depth frames using the method in Sec.~\ref{sec:yan_gen_mm2v}, and then distill it into a few-step casual model for real-time interaction through post-training in Sec.~\ref{sec:yan_gen_arpt} and Sec.~\ref{sec:yan_gen_sfpt}. Thanks to VACE~\citep{vace}, we find that directly combining the \texttt{Yan-Gen} model before post-training in Sec.~\ref{sec:yan_gen_mm2v} with the open-source ControlNet weights of VACE can perform versatile and temporally consistent style rendering on depth frames. 
Therefore, we only need to distill the combined model into the corresponding few-step causal model following the post-training method in \texttt{Yan-Gen}. To generate depth videos for distillation, we employ the trained interactive mechanics simulator to generate diverse depth videos by giving random input (i.e., random initial depths, random action signals, and random structure prompts). The length of each generated depth video is 81 frames. The style prompts used for distillation contain two parts: in-domain style prompts and out-of-domain style prompts. We extract the style description part from each caption generated in Sec .~\ref {sec:yan_gen_hier_caption} as the in-domain style prompt. The out-of-domain style prompts describing open-domain styles are generated by GPT-4~\citep{gpt4}, with the aim of improving the open-domain style rendering ability of the distillation model. 

After training, \texttt{Yan-Edit} is able to generate infinite-length videos auto-regressively in an interactive way. Specifically, given an initial depth frame, the interactive mechanics simulator receives the action signals and the structure prompts from the user to generate the next depth frames, which are then input into the visual render for style rendering conditioned on the style prompts to obtain the final result frames. Every generated frame will be stored in a KV cache for generating subsequent frames, which ensures the temporal consistency of the generated video. The detailed inference strategies of the interactive mechanics simulator and the visual renderer follow \texttt{Yan-Sim} and \texttt{Yan-Gen}, respectively. Through this frame-by-frame inference strategy, \texttt{Yan-Edit} enables infinite-length video generation conditioned on per-frame action signal of the user, structure prompt, and style prompt. More importantly, it allows users to input new structure prompts and style prompts at any moment during interaction to alter the structure and style of video content.

\subsubsection{Evaluation}
\label{yan_edit:eval}

In this subsection, we show qualitative results to verify the effectiveness of \texttt{Yan-Edit} for interactive video editing. Firstly, we demonstrate the structure editing abilities of \texttt{Yan-Edit}, which enables users to edit structures of the video content through structure prompts during interaction. Then, we present the style editing capabilities of \texttt{Yan-Edit} that can render the video content in diverse styles through style prompts. 

\begin{figure}[t]
\centering
\includegraphics[width=0.99\textwidth]{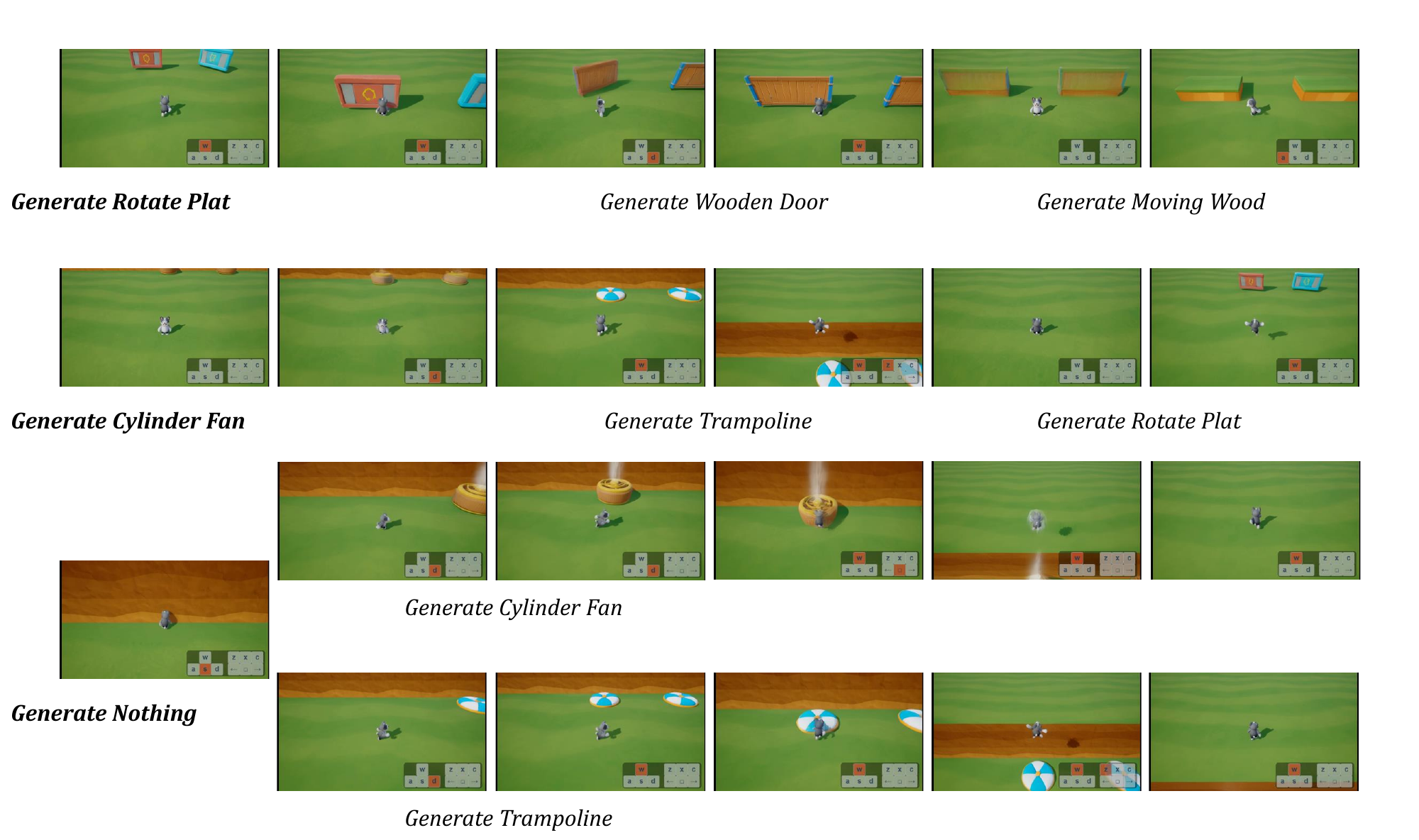}
\caption{
The visualization results of structure editing using \texttt{Yan-Edit}. We edit the scene structures during the interaction by replacing the scene elements with others. The user's per-frame interaction signal (i.e., keyboard control signal) is displayed in the lower right corner of each image.
}
\label{fig:structure_editing_results}
\end{figure}

\paragraph{Structure Editing.} Structure editing enables dynamic content creation and replacement within the scene environment. As shown in Fig.~\ref{fig:structure_editing_results}, through a structure prompt that switches the \textit{"Rotate Plat"} to \textit{"Wooden Door"}, the target element undergoes immediate transformation within the scene. Furthermore, when encountering obstructive structures such as walls, the system can generate elements in structure prompts like \textit{"Cylinder Fan"} or \textit{"Trampoline"} to resolve these obstacles. Consequently, \texttt{Yan-Edit} demonstrates capabilities not only in content creation but also in real-time element replacement.

\begin{figure}[t]
\centering
\includegraphics[width=0.99\textwidth]{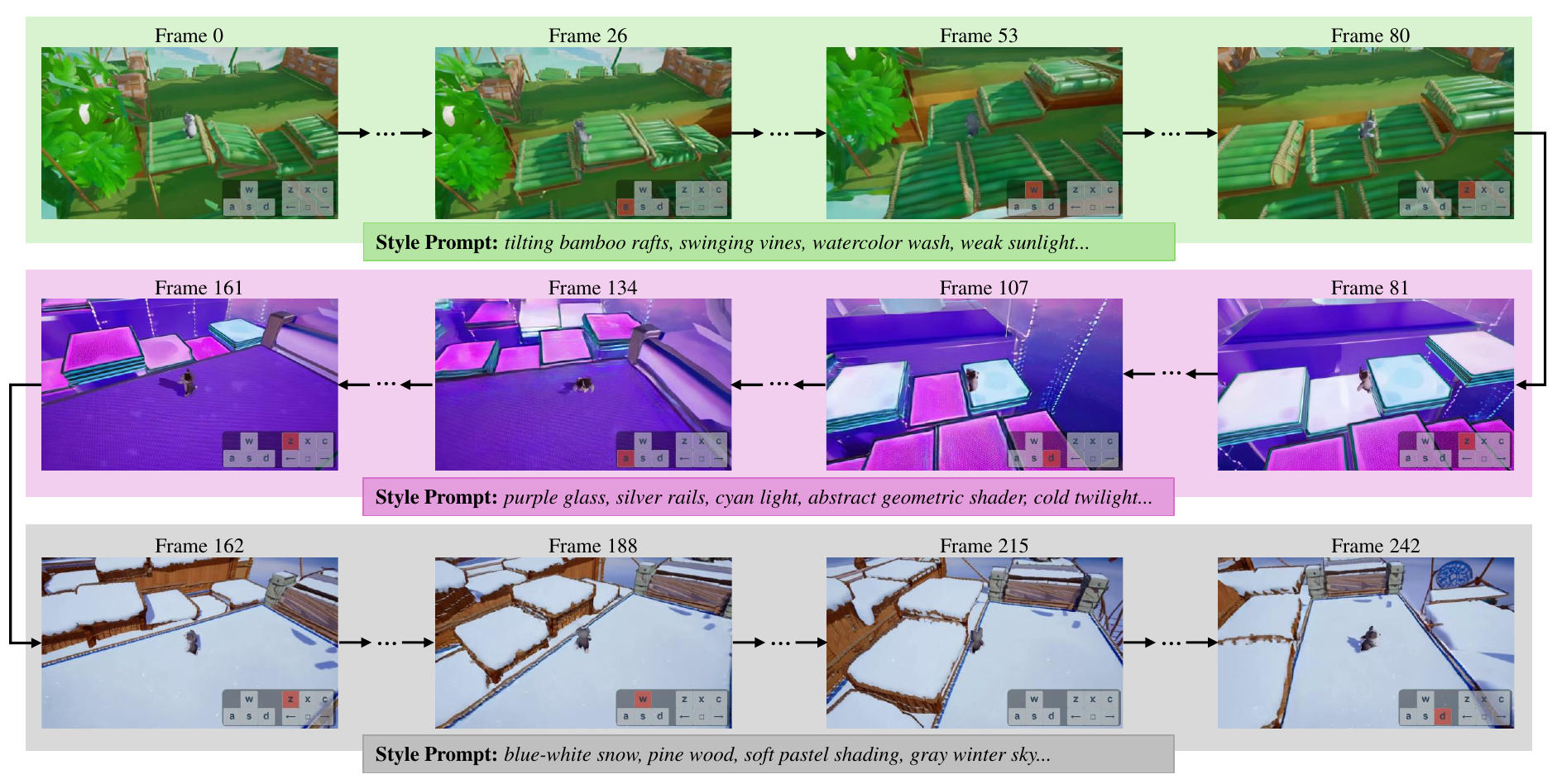}
\caption{
The visualization results of style editing using \texttt{Yan-Edit}. We edit the style during interaction by modifying the style prompt at frames 81 and 162. The user's per-frame interaction signal (i.e., keyboard control signal) is displayed in the lower right corner of each image.
}
\label{fig:style_editing_results}
\end{figure}

\paragraph{Style Editing.}
Style editing specifically refers to changing the color and texture of objects in the video. As shown in Fig.~\ref{fig:style_editing_results}, we use \textit{"bamboo and watercolor"} style to render the frame at the beginning of the video, and then switch the style to \textit{"purple glass and cyan light"} and \textit{"snow and soft pastel shading"} at frame 81 and frame 162, respectively. It can be observed that \texttt{Yan-Edit} is able to achieve versatile open-domain style editing with the visual renderer, while ensuring the accurate simulation of interactive mechanics through the interactive mechanics simulator. Besides, it allows users to seamlessly change the rendering style through style prompts at any moment during interaction.

\section{Limitation and Conclusion} 

\paragraph{Limitation}
We present \texttt{Yan}, a foundational interactive video generation framework achieving state-of-the-art capabilities in simulation, generation, and editing. Despite these advances, several limitations remain to be addressed to realize the full potential of interactive video generation. First, ensuring visual consistency across long video durations remains a challenge. Further improvements in spatial and temporal consistency, especially under complex and rapidly changing interactive scenarios, are warranted. Second, \texttt{Yan}'s high-fidelity, real-time performance is currently enabled by inference on powerful GPUs, which may limit accessibility for users with limited hardware resources. More efficient architectures and optimization for low-resource and edge devices are promising directions. Third, while \texttt{Yan} demonstrates strong generalizability to both in-domain and out-of-domain scenes, the action space and interaction complexity are still bounded by the underlying game environment, potentially restricting extensibility to some real-world applications. Finally, \texttt{Yan}'s editing module, though supporting both structure and style modifications in real time, currently relies on textual descriptions and may require further enhancement for intuitive and broader user control. 

\paragraph{Conclusion}
We introduced \texttt{Yan}, a foundational framework for interactive video generation that unifies (i) high-fidelity real-time simulation, (ii) prompt-controllable multi-modal generation, and (iii) multi-granularity, text-driven editing upon a large, precisely annotated dataset. Our modular yet coherent design bridges simulation, creation, and on-the-fly customization, demonstrating a practical path toward next-generation AI content engines. Future work will concentrate on (i) scaling both the dataset and model capacity while improving efficiency, (ii) extending prompt controllability to real-world domains, and (iii) developing richer evaluation protocols that account for human factors and safety.  We hope \texttt{Yan} serves as a stepping-stone toward more versatile, inclusive, and creative interactive media applications.

\section{Contributors}
We sincerely thank all contributors for their great efforts. The following contributors are listed in alphabetical order:

Deheng Ye, Fangyun Zhou, Jiacheng Lv, Jianqi Ma, Jun Zhang, Junyan Lv, Junyou Li, Minwen Deng, Mingyu Yang, Qiang Fu, Wei Yang, Wenkai Lv, Yangbin Yu, Yewen Wang, Yonghang Guan, Zhihao Hu, Zhongbin Fang, Zhongqian Sun.
\newpage
\bibliography{reference}

\begin{thebibliography}{61}
\providecommand{\natexlab}[1]{#1}
\providecommand{\url}[1]{\texttt{#1}}
\expandafter\ifx\csname urlstyle\endcsname\relax
  \providecommand{\doi}[1]{doi: #1}\else
  \providecommand{\doi}{doi: \begingroup \urlstyle{rm}\Url}\fi

\bibitem[Achiam et~al.(2023)Achiam, Adler, Agarwal, Ahmad, Akkaya, Aleman, Almeida, Altenschmidt, Altman, Anadkat, et~al.]{gpt4}
Josh Achiam, Steven Adler, Sandhini Agarwal, Lama Ahmad, Ilge Akkaya, Florencia~Leoni Aleman, Diogo Almeida, Janko Altenschmidt, Sam Altman, Shyamal Anadkat, et~al.
\newblock Gpt-4 technical report.
\newblock \emph{arXiv preprint arXiv:2303.08774}, 2023.

\bibitem[Agarwal et~al.(2025)Agarwal, Ali, Bala, Balaji, Barker, Cai, Chattopadhyay, Chen, Cui, Ding, et~al.]{cosmos}
Niket Agarwal, Arslan Ali, Maciej Bala, Yogesh Balaji, Erik Barker, Tiffany Cai, Prithvijit Chattopadhyay, Yongxin Chen, Yin Cui, Yifan Ding, et~al.
\newblock Cosmos world foundation model platform for physical ai.
\newblock \emph{arXiv preprint arXiv:2501.03575}, 2025.

\bibitem[Alonso et~al.(2024)Alonso, Jelley, Micheli, Kanervisto, Storkey, Pearce, and Fleuret]{diamond}
Eloi Alonso, Adam Jelley, Vincent Micheli, Anssi Kanervisto, Amos Storkey, Tim Pearce, and Fran{\c{c}}ois Fleuret.
\newblock Diffusion for world modeling: Visual details matter in atari.
\newblock \emph{arXiv preprint arXiv:2405.12399}, 2024.

\bibitem[Amidos(2019)]{java_mario}
Amidos.
\newblock Mario-ai-framework, 2019.
\newblock URL \url{https://github.com/amidos2006/Mario-AI-Framework}.

\bibitem[Bai et~al.(2025{\natexlab{a}})Bai, Chen, Liu, Wang, Ge, Song, Dang, Wang, Wang, Tang, Zhong, Zhu, Yang, Li, Wan, Wang, Ding, Fu, Xu, Ye, Zhang, Xie, Cheng, Zhang, Yang, Xu, and Lin]{bai2025qwen25vltechnicalreport}
Shuai Bai, Keqin Chen, Xuejing Liu, Jialin Wang, Wenbin Ge, Sibo Song, Kai Dang, Peng Wang, Shijie Wang, Jun Tang, Humen Zhong, Yuanzhi Zhu, Mingkun Yang, Zhaohai Li, Jianqiang Wan, Pengfei Wang, Wei Ding, Zheren Fu, Yiheng Xu, Jiabo Ye, Xi~Zhang, Tianbao Xie, Zesen Cheng, Hang Zhang, Zhibo Yang, Haiyang Xu, and Junyang Lin.
\newblock Qwen2.5-vl technical report, 2025{\natexlab{a}}.
\newblock URL \url{https://arxiv.org/abs/2502.13923}.

\bibitem[Bai et~al.(2025{\natexlab{b}})Bai, Chen, Liu, Wang, Ge, Song, Dang, Wang, Wang, Tang, et~al.]{bai2025qwen2}
Shuai Bai, Keqin Chen, Xuejing Liu, Jialin Wang, Wenbin Ge, Sibo Song, Kai Dang, Peng Wang, Shijie Wang, Jun Tang, et~al.
\newblock Qwen2. 5-vl technical report.
\newblock \emph{arXiv preprint arXiv:2502.13923}, 2025{\natexlab{b}}.

\bibitem[Ball et~al.(2025)Ball, Bauer, Belletti, Brownfield, Ephrat, Fruchter, Gupta, Holsheimer, Holynski, Hron, Kaplanis, Limont, McGill, Oliveira, Parker-Holder, Perbet, Scully, Shar, Spencer, Tov, Villegas, Wang, Yung, Baetu, Berbel, Bridson, Bruce, Buttimore, Chakera, Chandra, Collins, Cullum, Damoc, Dasagi, Gazeau, Gbadamosi, Han, Hirst, Kachra, Kerley, Kjems, Knoepfel, Koriakin, Lo, Lu, Mehring, Moufarek, Nandwani, Oliveira, Pardo, Park, Pierson, Poole, Ran, Salimans, Sanchez, Saprykin, Shen, Sidhwani, Smith, Stanton, Tomlinson, Vijaykumar, Wang, Wingfield, Wong, Xu, Yew, Young, Zubov, Eck, Erhan, Kavukcuoglu, Hassabis, Gharamani, Hadsell, van~den Oord, Mosseri, Bolton, Singh, and Rockt{\"a}schel]{genie3}
Philip~J. Ball, Jakob Bauer, Frank Belletti, Bethanie Brownfield, Ariel Ephrat, Shlomi Fruchter, Agrim Gupta, Kristian Holsheimer, Aleksander Holynski, Jiri Hron, Christos Kaplanis, Marjorie Limont, Matt McGill, Yanko Oliveira, Jack Parker-Holder, Frank Perbet, Guy Scully, Jeremy Shar, Stephen Spencer, Omer Tov, Ruben Villegas, Emma Wang, Jessica Yung, Cip Baetu, Jordi Berbel, David Bridson, Jake Bruce, Gavin Buttimore, Sarah Chakera, Bilva Chandra, Paul Collins, Alex Cullum, Bogdan Damoc, Vibha Dasagi, Maxime Gazeau, Charles Gbadamosi, Woohyun Han, Ed~Hirst, Ashyana Kachra, Lucie Kerley, Kristian Kjems, Eva Knoepfel, Vika Koriakin, Jessica Lo, Cong Lu, Zeb Mehring, Alex Moufarek, Henna Nandwani, Valeria Oliveira, Fabio Pardo, Jane Park, Andrew Pierson, Ben Poole, Helen Ran, Tim Salimans, Manuel Sanchez, Igor Saprykin, Amy Shen, Sailesh Sidhwani, Duncan Smith, Joe Stanton, Hamish Tomlinson, Dimple Vijaykumar, Luyu Wang, Piers Wingfield, Nat Wong, Keyang Xu, Christopher Yew, Nick Young, Vadim Zubov, Douglas
  Eck, Dumitru Erhan, Koray Kavukcuoglu, Demis Hassabis, Zoubin Gharamani, Raia Hadsell, A{\"a}ron van~den Oord, Inbar Mosseri, Adrian Bolton, Satinder Singh, and Tim Rockt{\"a}schel.
\newblock Genie 3: A new frontier for world models, 2025.
\newblock URL \url{https://deepmind.google/discover/blog/genie-3-a-new-frontier-for-world-models/}.

\bibitem[Bar et~al.(2025)Bar, Zhou, Tran, Darrell, and LeCun]{nwm}
Amir Bar, Gaoyue Zhou, Danny Tran, Trevor Darrell, and Yann LeCun.
\newblock Navigation world models.
\newblock In \emph{Proceedings of the Computer Vision and Pattern Recognition Conference}, pp.\  15791--15801, 2025.

\bibitem[Brooks et~al.(2024)Brooks, Peebles, Holmes, DePue, Guo, Jing, Schnurr, Taylor, Luhman, Luhman, Ng, Wang, and Ramesh]{sora}
Tim Brooks, Bill Peebles, Connor Holmes, Will DePue, Yufei Guo, Li~Jing, David Schnurr, Joe Taylor, Troy Luhman, Eric Luhman, Clarence Ng, Ricky Wang, and Aditya Ramesh.
\newblock Video generation models as world simulators, 2024.
\newblock URL \url{https://openai.com/research/video-generation-models-as-world-simulators}.

\bibitem[Bruce et~al.(2024)Bruce, Dennis, Edwards, Parker-Holder, Shi, Hughes, Lai, Mavalankar, Steigerwald, Apps, et~al.]{genie}
Jake Bruce, Michael~D Dennis, Ashley Edwards, Jack Parker-Holder, Yuge Shi, Edward Hughes, Matthew Lai, Aditi Mavalankar, Richie Steigerwald, Chris Apps, et~al.
\newblock Genie: Generative interactive environments.
\newblock In \emph{Forty-first International Conference on Machine Learning}, 2024.

\bibitem[Ceylan et~al.(2023)Ceylan, Huang, and Mitra]{pix2video}
Duygu Ceylan, Chun-Hao~P Huang, and Niloy~J Mitra.
\newblock Pix2video: Video editing using image diffusion.
\newblock In \emph{Proceedings of the IEEE/CVF International Conference on Computer Vision}, pp.\  23206--23217, 2023.

\bibitem[Che et~al.(2024)Che, He, Liu, Jin, and Chen]{gamegen_x}
Haoxuan Che, Xuanhua He, Quande Liu, Cheng Jin, and Hao Chen.
\newblock Gamegen-x: Interactive open-world game video generation.
\newblock \emph{arXiv preprint arXiv:2411.00769}, 2024.

\bibitem[Chen et~al.(2024{\natexlab{a}})Chen, Mart{\'\i}~Mons{\'o}, Du, Simchowitz, Tedrake, and Sitzmann]{chen2024diffusion}
Boyuan Chen, Diego Mart{\'\i}~Mons{\'o}, Yilun Du, Max Simchowitz, Russ Tedrake, and Vincent Sitzmann.
\newblock Diffusion forcing: Next-token prediction meets full-sequence diffusion.
\newblock \emph{Advances in Neural Information Processing Systems}, 37:\penalty0 24081--24125, 2024{\natexlab{a}}.

\bibitem[Chen et~al.(2025)Chen, Ge, Zhang, Zhang, Zhu, Yang, Hao, Wu, Lai, Hu, et~al.]{goku}
Shoufa Chen, Chongjian Ge, Yuqi Zhang, Yida Zhang, Fengda Zhu, Hao Yang, Hongxiang Hao, Hui Wu, Zhichao Lai, Yifei Hu, et~al.
\newblock Goku: Flow based video generative foundation models.
\newblock In \emph{Proceedings of the Computer Vision and Pattern Recognition Conference}, pp.\  23516--23527, 2025.

\bibitem[Chen et~al.(2024{\natexlab{b}})Chen, Siarohin, Menapace, Deyneka, Chao, Jeon, Fang, Lee, Ren, Yang, et~al.]{panda70m}
Tsai-Shien Chen, Aliaksandr Siarohin, Willi Menapace, Ekaterina Deyneka, Hsiang-wei Chao, Byung~Eun Jeon, Yuwei Fang, Hsin-Ying Lee, Jian Ren, Ming-Hsuan Yang, et~al.
\newblock Panda-70m: Captioning 70m videos with multiple cross-modality teachers.
\newblock In \emph{Proceedings of the IEEE/CVF Conference on Computer Vision and Pattern Recognition}, pp.\  13320--13331, 2024{\natexlab{b}}.

\bibitem[Chung et~al.(2023)Chung, Constant, Garcia, Roberts, Tay, Narang, and Firat]{chung2023unimax}
Hyung~Won Chung, Noah Constant, Xavier Garcia, Adam Roberts, Yi~Tay, Sharan Narang, and Orhan Firat.
\newblock Unimax: Fairer and more effective language sampling for large-scale multilingual pretraining.
\newblock \emph{arXiv preprint arXiv:2304.09151}, 2023.

\bibitem[Decart et~al.(2024)Decart, Julian, Quinn, Spruce, Xinlei, and Robert]{oasis}
Decart, Quevedo Julian, McIntyre Quinn, Campbell Spruce, Chen Xinlei, and Wachen Robert.
\newblock Oasis: A universe in a transformer, 2024.
\newblock URL \url{https://oasis-model.github.io/}.

\bibitem[Esser et~al.(2024)Esser, Kulal, Blattmann, Entezari, M{\"u}ller, Saini, Levi, Lorenz, Sauer, Boesel, et~al.]{esser2024scaling}
Patrick Esser, Sumith Kulal, Andreas Blattmann, Rahim Entezari, Jonas M{\"u}ller, Harry Saini, Yam Levi, Dominik Lorenz, Axel Sauer, Frederic Boesel, et~al.
\newblock Scaling rectified flow transformers for high-resolution image synthesis.
\newblock In \emph{Forty-first international conference on machine learning}, 2024.

\bibitem[Feng et~al.(2024)Feng, Zhang, Yang, Xiao, Shu, Liu, Zheng, Huang, Liu, and Zhang]{the_matrix}
Ruili Feng, Han Zhang, Zhantao Yang, Jie Xiao, Zhilei Shu, Zhiheng Liu, Andy Zheng, Yukun Huang, Yu~Liu, and Hongyang Zhang.
\newblock The matrix: Infinite-horizon world generation with real-time moving control.
\newblock \emph{arXiv preprint arXiv:2412.03568}, 2024.

\bibitem[Gui et~al.(2025)Gui, Schusterbauer, Prestel, Ma, Kotovenko, Grebenkova, Baumann, Hu, and Ommer]{depthfm}
Ming Gui, Johannes Schusterbauer, Ulrich Prestel, Pingchuan Ma, Dmytro Kotovenko, Olga Grebenkova, Stefan~Andreas Baumann, Vincent~Tao Hu, and Bj{\"o}rn Ommer.
\newblock Depthfm: Fast generative monocular depth estimation with flow matching.
\newblock In \emph{Proceedings of the AAAI Conference on Artificial Intelligence}, pp.\  3203--3211, 2025.

\bibitem[Guo et~al.(2025)Guo, Ye, He, Wu, Jiang, Pearce, and Bian]{mineworld}
Junliang Guo, Yang Ye, Tianyu He, Haoyu Wu, Yushu Jiang, Tim Pearce, and Jiang Bian.
\newblock Mineworld: a real-time and open-source interactive world model on minecraft.
\newblock \emph{arXiv preprint arXiv:2504.08388}, 2025.

\bibitem[Guo et~al.(2023)Guo, Yang, Rao, Liang, Wang, Qiao, Agrawala, Lin, and Dai]{guo2023animatediff}
Yuwei Guo, Ceyuan Yang, Anyi Rao, Zhengyang Liang, Yaohui Wang, Yu~Qiao, Maneesh Agrawala, Dahua Lin, and Bo~Dai.
\newblock Animatediff: Animate your personalized text-to-image diffusion models without specific tuning.
\newblock \emph{arXiv preprint arXiv:2307.04725}, 2023.

\bibitem[Ho et~al.(2020)Ho, Jain, and Abbeel]{ho2020denoising}
Jonathan Ho, Ajay Jain, and Pieter Abbeel.
\newblock Denoising diffusion probabilistic models.
\newblock \emph{Advances in neural information processing systems}, 33:\penalty0 6840--6851, 2020.

\bibitem[Huang et~al.(2025)Huang, Li, He, Zhou, and Shechtman]{huang2025self}
Xun Huang, Zhengqi Li, Guande He, Mingyuan Zhou, and Eli Shechtman.
\newblock Self forcing: Bridging the train-test gap in autoregressive video diffusion.
\newblock \emph{arXiv preprint arXiv:2506.08009}, 2025.

\bibitem[Jiang et~al.(2025)Jiang, Han, Mao, Zhang, Pan, and Liu]{vace}
Zeyinzi Jiang, Zhen Han, Chaojie Mao, Jingfeng Zhang, Yulin Pan, and Yu~Liu.
\newblock Vace: All-in-one video creation and editing.
\newblock \emph{arXiv preprint arXiv:2503.07598}, 2025.

\bibitem[Kempka et~al.(2016)Kempka, Wydmuch, Runc, Toczek, and Ja{\'s}kowski]{vizdoom}
Micha{\l} Kempka, Marek Wydmuch, Grzegorz Runc, Jakub Toczek, and Wojciech Ja{\'s}kowski.
\newblock Vizdoom: A doom-based ai research platform for visual reinforcement learning.
\newblock In \emph{2016 IEEE conference on computational intelligence and games (CIG)}, pp.\  1--8. IEEE, 2016.

\bibitem[Kingma et~al.(2013)Kingma, Welling, et~al.]{kingma2013auto}
Diederik~P Kingma, Max Welling, et~al.
\newblock Auto-encoding variational bayes, 2013.

\bibitem[Kong et~al.(2024)Kong, Tian, Zhang, Min, Dai, Zhou, Xiong, Li, Wu, Zhang, et~al.]{hunyuanvideo}
Weijie Kong, Qi~Tian, Zijian Zhang, Rox Min, Zuozhuo Dai, Jin Zhou, Jiangfeng Xiong, Xin Li, Bo~Wu, Jianwei Zhang, et~al.
\newblock Hunyuanvideo: A systematic framework for large video generative models.
\newblock \emph{arXiv preprint arXiv:2412.03603}, 2024.

\bibitem[Li et~al.(2024)Li, Zhang, Yu, Fu, and Ye]{ma}
Junyou Li, Qin Zhang, Yangbin Yu, Qiang Fu, and Deheng Ye.
\newblock More agents is all you need.
\newblock \emph{arXiv preprint arXiv:2402.05120}, 2024.

\bibitem[Liu et~al.(2024)Liu, Zhang, Li, Lin, and Jia]{video_p2p}
Shaoteng Liu, Yuechen Zhang, Wenbo Li, Zhe Lin, and Jiaya Jia.
\newblock Video-p2p: Video editing with cross-attention control.
\newblock In \emph{Proceedings of the IEEE/CVF Conference on Computer Vision and Pattern Recognition}, pp.\  8599--8608, 2024.

\bibitem[Nan et~al.(2024)Nan, Xie, Zhou, Fan, Yang, Chen, Li, Yang, and Tai]{openvid-1m}
Kepan Nan, Rui Xie, Penghao Zhou, Tiehan Fan, Zhenheng Yang, Zhijie Chen, Xiang Li, Jian Yang, and Ying Tai.
\newblock Openvid-1m: A large-scale high-quality dataset for text-to-video generation.
\newblock \emph{arXiv preprint arXiv:2407.02371}, 2024.

\bibitem[Parker-Holder et~al.(2024)Parker-Holder, Ball, Bruce, Dasagi, Holsheimer, Kaplanis, Moufarek, Scully, Shar, Shi, Spencer, Yung, Dennis, Kenjeyev, Long, Mnih, Chan, Gazeau, Li, Pardo, Wang, Zhang, Besse, Harley, Mitenkova, Wang, Clune, Hassabis, Hadsell, Bolton, Singh, and Rocktäschel]{genie2}
Jack Parker-Holder, Philip Ball, Jake Bruce, Vibhavari Dasagi, Kristian Holsheimer, Christos Kaplanis, Alexandre Moufarek, Guy Scully, Jeremy Shar, Jimmy Shi, Stephen Spencer, Jessica Yung, Michael Dennis, Sultan Kenjeyev, Shangbang Long, Vlad Mnih, Harris Chan, Maxime Gazeau, Bonnie Li, Fabio Pardo, Luyu Wang, Lei Zhang, Frederic Besse, Tim Harley, Anna Mitenkova, Jane Wang, Jeff Clune, Demis Hassabis, Raia Hadsell, Adrian Bolton, Satinder Singh, and Tim Rocktäschel.
\newblock Genie 2: A large-scale foundation world model, 2024.
\newblock URL \url{https://deepmind.google/discover/blog/genie-2-a-large-scale-foundation-world-model/}.

\bibitem[Pinto(2021)]{mario_dataset}
R.C. Pinto.
\newblock Super mario bros. gameplay dataset.
\newblock \url{https://github.com/rafaelcp/smbdataset}, 2021.

\bibitem[Protocol(2024)]{mariovgg}
Virtuals Protocol.
\newblock Video game generation: A practical study using mario, 2024.
\newblock URL \url{https://github.com/Virtual-Protocol/mario-videogamegen/blob/main/static/pdfs/VideoGameGen.pdf}.
\newblock Preprint.

\bibitem[Radford et~al.(2021)Radford, Kim, Hallacy, Ramesh, Goh, Agarwal, Sastry, Askell, Mishkin, Clark, et~al.]{radford2021learning}
Alec Radford, Jong~Wook Kim, Chris Hallacy, Aditya Ramesh, Gabriel Goh, Sandhini Agarwal, Girish Sastry, Amanda Askell, Pamela Mishkin, Jack Clark, et~al.
\newblock Learning transferable visual models from natural language supervision.
\newblock In \emph{International conference on machine learning}, pp.\  8748--8763. PmLR, 2021.

\bibitem[Rombach et~al.(2022{\natexlab{a}})Rombach, Blattmann, Lorenz, Esser, and Ommer]{rombach2022high}
Robin Rombach, Andreas Blattmann, Dominik Lorenz, Patrick Esser, and Bj{\"o}rn Ommer.
\newblock High-resolution image synthesis with latent diffusion models.
\newblock In \emph{Proceedings of the IEEE/CVF conference on computer vision and pattern recognition}, pp.\  10684--10695, 2022{\natexlab{a}}.

\bibitem[Rombach et~al.(2022{\natexlab{b}})Rombach, Blattmann, Lorenz, Esser, and Ommer]{stable_diffusion}
Robin Rombach, Andreas Blattmann, Dominik Lorenz, Patrick Esser, and Bj{\"o}rn Ommer.
\newblock High-resolution image synthesis with latent diffusion models.
\newblock In \emph{Proceedings of the IEEE/CVF conference on computer vision and pattern recognition}, pp.\  10684--10695, 2022{\natexlab{b}}.

\bibitem[Schulman et~al.(2017)Schulman, Wolski, Dhariwal, Radford, and Klimov]{ppo}
John Schulman, Filip Wolski, Prafulla Dhariwal, Alec Radford, and Oleg Klimov.
\newblock Proximal policy optimization algorithms.
\newblock \emph{arXiv preprint arXiv:1707.06347}, 2017.

\bibitem[Shi et~al.(2016)Shi, Caballero, Husz{\'a}r, Totz, Aitken, Bishop, Rueckert, and Wang]{shi2016real}
Wenzhe Shi, Jose Caballero, Ferenc Husz{\'a}r, Johannes Totz, Andrew~P Aitken, Rob Bishop, Daniel Rueckert, and Zehan Wang.
\newblock Real-time single image and video super-resolution using an efficient sub-pixel convolutional neural network.
\newblock In \emph{Proceedings of the IEEE conference on computer vision and pattern recognition}, pp.\  1874--1883, 2016.

\bibitem[Song et~al.(2020)Song, Meng, and Ermon]{song2020denoising}
Jiaming Song, Chenlin Meng, and Stefano Ermon.
\newblock Denoising diffusion implicit models.
\newblock \emph{arXiv preprint arXiv:2010.02502}, 2020.

\bibitem[Song et~al.(2025)Song, Chen, Simchowitz, Du, Tedrake, and Sitzmann]{song2025history}
Kiwhan Song, Boyuan Chen, Max Simchowitz, Yilun Du, Russ Tedrake, and Vincent Sitzmann.
\newblock History-guided video diffusion.
\newblock \emph{arXiv preprint arXiv:2502.06764}, 2025.

\bibitem[Studios(2011)]{minecraft}
Mojang Studios.
\newblock Minecraft.
\newblock \url{https://www.minecraft.net/}, 2011.
\newblock Video game.

\bibitem[Tencent(2023)]{ymzx}
Tencent.
\newblock Yuan meng star.
\newblock \url{https://en.wikipedia.org/wiki/Yuan_Meng_Star}, 2023.

\bibitem[Valevski et~al.(2024)Valevski, Leviathan, Arar, and Fruchter]{gameNgen}
Dani Valevski, Yaniv Leviathan, Moab Arar, and Shlomi Fruchter.
\newblock Diffusion models are real-time game engines.
\newblock \emph{arXiv preprint arXiv:2408.14837}, 2024.

\bibitem[Wan et~al.(2025{\natexlab{a}})Wan, Wang, Ai, Wen, Mao, Xie, Chen, Yu, Zhao, Yang, et~al.]{wan}
Team Wan, Ang Wang, Baole Ai, Bin Wen, Chaojie Mao, Chen-Wei Xie, Di~Chen, Feiwu Yu, Haiming Zhao, Jianxiao Yang, et~al.
\newblock Wan: Open and advanced large-scale video generative models.
\newblock \emph{arXiv preprint arXiv:2503.20314}, 2025{\natexlab{a}}.

\bibitem[Wan et~al.(2025{\natexlab{b}})Wan, Wang, Ai, Wen, Mao, Xie, Chen, Yu, Zhao, Yang, et~al.]{wan2025wan}
Team Wan, Ang Wang, Baole Ai, Bin Wen, Chaojie Mao, Chen-Wei Xie, Di~Chen, Feiwu Yu, Haiming Zhao, Jianxiao Yang, et~al.
\newblock Wan: Open and advanced large-scale video generative models.
\newblock \emph{arXiv preprint arXiv:2503.20314}, 2025{\natexlab{b}}.

\bibitem[Wang et~al.(2023{\natexlab{a}})Wang, Yuan, Chen, Zhang, Wang, and Zhang]{wang2023modelscope}
Jiuniu Wang, Hangjie Yuan, Dayou Chen, Yingya Zhang, Xiang Wang, and Shiwei Zhang.
\newblock Modelscope text-to-video technical report.
\newblock \emph{arXiv preprint arXiv:2308.06571}, 2023{\natexlab{a}}.

\bibitem[Wang et~al.(2023{\natexlab{b}})Wang, He, Li, Li, Yu, Ma, Li, Chen, Chen, Wang, et~al.]{internvid}
Yi~Wang, Yinan He, Yizhuo Li, Kunchang Li, Jiashuo Yu, Xin Ma, Xinhao Li, Guo Chen, Xinyuan Chen, Yaohui Wang, et~al.
\newblock Internvid: A large-scale video-text dataset for multimodal understanding and generation.
\newblock \emph{arXiv preprint arXiv:2307.06942}, 2023{\natexlab{b}}.

\bibitem[Yang et~al.(2023)Yang, Du, Ghasemipour, Tompson, Schuurmans, and Abbeel]{unisim}
Mengjiao Yang, Yilun Du, Kamyar Ghasemipour, Jonathan Tompson, Dale Schuurmans, and Pieter Abbeel.
\newblock Learning interactive real-world simulators.
\newblock \emph{arXiv preprint arXiv:2310.06114}, 1\penalty0 (2):\penalty0 6, 2023.

\bibitem[Yang et~al.(2024)Yang, Li, Fang, Chen, Yu, Fu, Yang, and Ye]{playgen}
Mingyu Yang, Junyou Li, Zhongbin Fang, Sheng Chen, Yangbin Yu, Qiang Fu, Wei Yang, and Deheng Ye.
\newblock Playable game generation.
\newblock \emph{arXiv preprint arXiv:2412.00887}, 2024.

\bibitem[Yin et~al.(2024{\natexlab{a}})Yin, Gharbi, Park, Zhang, Shechtman, Durand, and Freeman]{yin2024improved}
Tianwei Yin, Micha{\"e}l Gharbi, Taesung Park, Richard Zhang, Eli Shechtman, Fredo Durand, and Bill Freeman.
\newblock Improved distribution matching distillation for fast image synthesis.
\newblock \emph{Advances in neural information processing systems}, 37:\penalty0 47455--47487, 2024{\natexlab{a}}.

\bibitem[Yin et~al.(2024{\natexlab{b}})Yin, Gharbi, Zhang, Shechtman, Durand, Freeman, and Park]{yin2024one}
Tianwei Yin, Micha{\"e}l Gharbi, Richard Zhang, Eli Shechtman, Fredo Durand, William~T Freeman, and Taesung Park.
\newblock One-step diffusion with distribution matching distillation.
\newblock In \emph{Proceedings of the IEEE/CVF conference on computer vision and pattern recognition}, pp.\  6613--6623, 2024{\natexlab{b}}.

\bibitem[Yin et~al.(2025)Yin, Zhang, Zhang, Freeman, Durand, Shechtman, and Huang]{yin2025slow}
Tianwei Yin, Qiang Zhang, Richard Zhang, William~T Freeman, Fredo Durand, Eli Shechtman, and Xun Huang.
\newblock From slow bidirectional to fast autoregressive video diffusion models.
\newblock In \emph{Proceedings of the Computer Vision and Pattern Recognition Conference}, pp.\  22963--22974, 2025.

\bibitem[Yu et~al.(2025{\natexlab{a}})Yu, Qin, Che, Liu, Wang, Wan, Zhang, Gai, Chen, and Liu]{igv_survey}
Jiwen Yu, Yiran Qin, Haoxuan Che, Quande Liu, Xintao Wang, Pengfei Wan, Di~Zhang, Kun Gai, Hao Chen, and Xihui Liu.
\newblock A survey of interactive generative video.
\newblock \emph{arXiv preprint arXiv:2504.21853}, 2025{\natexlab{a}}.

\bibitem[Yu et~al.(2025{\natexlab{b}})Yu, Qin, Wang, Wan, Zhang, and Liu]{gamefactory}
Jiwen Yu, Yiran Qin, Xintao Wang, Pengfei Wan, Di~Zhang, and Xihui Liu.
\newblock Gamefactory: Creating new games with generative interactive videos.
\newblock \emph{arXiv preprint arXiv:2501.08325}, 2025{\natexlab{b}}.

\bibitem[Yu et~al.(2024)Yu, Zhang, Li, Fu, and Ye]{aga}
Yangbin Yu, Qin Zhang, Junyou Li, Qiang Fu, and Deheng Ye.
\newblock Affordable generative agents.
\newblock \emph{arXiv preprint arXiv:2402.02053}, 2024.

\bibitem[Zhang et~al.(2023)Zhang, Rao, and Agrawala]{controlnet}
Lvmin Zhang, Anyi Rao, and Maneesh Agrawala.
\newblock Adding conditional control to text-to-image diffusion models.
\newblock In \emph{Proceedings of the IEEE/CVF international conference on computer vision}, pp.\  3836--3847, 2023.

\bibitem[Zhang et~al.(2018)Zhang, Isola, Efros, Shechtman, and Wang]{zhang2018unreasonable}
Richard Zhang, Phillip Isola, Alexei~A Efros, Eli Shechtman, and Oliver Wang.
\newblock The unreasonable effectiveness of deep features as a perceptual metric.
\newblock In \emph{Proceedings of the IEEE conference on computer vision and pattern recognition}, pp.\  586--595, 2018.

\bibitem[Zhang et~al.(2025)Zhang, Peng, Wang, Wang, Zhu, Gao, Li, Liu, and Zhou]{matrix_game}
Yifan Zhang, Chunli Peng, Boyang Wang, Puyi Wang, Qingcheng Zhu, Zedong Gao, Eric Li, Yang Liu, and Yahui Zhou.
\newblock Matrix-game: Interactive world foundation model.
\newblock \emph{arXiv}, 2025.

\bibitem[Zhao et~al.(2023)Zhao, Zhou, Li, Tang, Wang, Hou, Min, Zhang, Zhang, Dong, et~al.]{llm_survey}
Wayne~Xin Zhao, Kun Zhou, Junyi Li, Tianyi Tang, Xiaolei Wang, Yupeng Hou, Yingqian Min, Beichen Zhang, Junjie Zhang, Zican Dong, et~al.
\newblock A survey of large language models.
\newblock \emph{arXiv preprint arXiv:2303.18223}, 2023.

\bibitem[Zhou et~al.(2022)Zhou, Wang, Yan, Lv, Zhu, and Feng]{zhou2022magicvideo}
Daquan Zhou, Weimin Wang, Hanshu Yan, Weiwei Lv, Yizhe Zhu, and Jiashi Feng.
\newblock Magicvideo: Efficient video generation with latent diffusion models.
\newblock \emph{arXiv preprint arXiv:2211.11018}, 2022.

\end{thebibliography}
\bibliographystyle{tmlr}
\newpage
\appendix

\end{document}